%% file: neurips_2026_acknowledgement.tex
\documentclass{article}

    \PassOptionsToPackage{numbers, compress}{natbib}

\usepackage[preprint]{neurips_2026}

\usepackage[utf8]{inputenc} 
\usepackage[T1]{fontenc}    
\usepackage{hyperref}       
\usepackage{url}            
\usepackage{booktabs}       
\usepackage{amsfonts}       
\usepackage{nicefrac}       
\usepackage{microtype}      
\usepackage{xcolor}         
\usepackage{graphicx}
\usepackage{booktabs}
\usepackage{graphicx}
\usepackage{siunitx}

\usepackage{amsmath}
\usepackage{amssymb}
\usepackage{mathtools}
\usepackage{amsthm}
\usepackage{algorithm}
\usepackage{algorithmic}
\usepackage{caption}
\usepackage{subcaption}
\usepackage{todonotes}
\usepackage{wrapfig}
\usepackage{adjustbox}

\theoremstyle{plain}
\newtheorem{theorem}{Theorem}[section]

\theoremstyle{definition}
\newtheorem{definition}[theorem]{Definition}

\theoremstyle{remark}

\usepackage{subcaption}
\allowdisplaybreaks

\usepackage[capitalize,noabbrev]{cleveref}

\title{POP: Prior-Fitted First-Order Optimization Policies}

\author{%
  Jan Kobiolka\thanks{Corresponding author: \texttt{jan.kobiolka@utn.de}} \\
  Department of Computer Science\\
  University of Technology Nuremberg\\
  Germany \\
  \And
  Christian Frey \\
  Department of Computer Science\\
  University of Technology Nuremberg\\
  Germany\\
  \And
  Gresa Shala \\
  Department of Computer Science\\
  Albert Ludwig University of Freiburg\\
  Germany\\
  \And
  Arlind Kadra \\
  Department of Computer Science\\
  University of Technology Nuremberg\\
  Germany\\
  \And
  Erind Bedalli \\
  Faculty of Natural Sciences\\
  University of Elbasan\\
  Albania\\
  \And
  Josif Grabocka \\
  Department of Computer Science\\
  University of Technology Nuremberg\\
  Germany\\
}

\begin{document}

\maketitle

\input{00_abstract}
\input{01_introduction}
\input{03_methodology}

\input{04_prior}

\input{05_experimental_protocol}

\input{06_evaluation}
\input{02_related_work}

\input{07_conclusion}

\section*{Acknowledgments}
Josif Grabocka and Arlind Kadra acknowledge the funding support from the ”Bayerisches Landesamt fur Steuer” for the Bavarian AI Taxation Laboratory.

Gresa Shala and Josif Grabocka acknowledge the funding by The Carl Zeiss Foundation through the research network ”Responsive and Scalable Learning for Robots Assisting Humans” (ReScaLe) of the University of Freiburg

Erind Bedalli and Josif Grabocka acknowledge the financial support from the Albanian-American Development Foundation through the READ program.

Moreover, we gratefully acknowledge the scientific support and HPC resources provided by the Erlangen National High Performance Computing Center (NHR@FAU) of the Friedrich-Alexander-Universität Erlangen-Nürnberg (FAU). NHR@FAU hardware is partially funded by the German Research Foundation (DFG) – 440719683.


\newpage

\bibliographystyle{splncs04}
\bibliography{references}


\newpage
\appendix

\input{08_gp_proof}
\input{09_appendix_b}
\input{09_appendix_c}
\input{10_appendix_d}
\input{10_appendix_h}



\end{document}

%% file: 00_abstract.tex
\begin{abstract}

Gradient-based optimizers are highly sensitive to design choices in their adaptive learning rate mechanisms. To address this limitation, we introduce POP, a meta-learned Reinforcement Learning (RL) policy that predicts adaptive learning rates for gradient descent, conditioned on the contextual information provided in the optimization trajectory. Our method introduces a novel RL reward formulation, a new function-scaling strategy for in-distribution generalization, and a novel prior that is used to sample millions of synthetic optimization problems. We evaluate POP on an established benchmark including 43 optimization functions of various complexity, where it significantly outperforms gradient-based methods. Our evaluation demonstrates strong generalization capabilities without task-specific tuning.
\end{abstract}

%% file: 01_introduction.tex
\section{Introduction}
\label{sec:introduction}
Optimization strategies aim to identify the best possible solution by maximizing desired outcomes within a set of defined constraints. 
In practice, most continuous optimization problems are solved using gradient and momentum-based methods 
focusing on fast, robust convergence~
\citep{RobbinsMonro_1951_SGD,Kingma_2014_adam,Daoud_2022_GBO,Li_2020_SGDAnalysis}. Despite their success, the performance of these optimizers is highly sensitive to hyperparameter choices, in particular the learning-rate schedule and momentum coefficients \citep{Tian_2023_Recent,Keskar_2017_Improving,Hassan_2022_ChooingOptCV}. Selecting suitable hyperparameters often requires extensive hyperparameter optimization, which is costly, time-consuming, and difficult to transfer across tasks and domains \cite{Mary_2025_Optimizing,Liao_2022_HPOImpact,Schlotthauer_2025_Pre-Training}.

More recently, learned optimizers have gained increasing interest in the community \cite{Tang_2024_L20Overview,Yang_2023_LearningGenL20,Metz_2022_VeLO,Cao_2019_LearningSwarms,Gomes_2021_Meta_BBPBO,Vishnu_2020_MetaLearningBBO,Ma_2024_HeapBasedOpt,Goldie_2024_LearningOptRL,Lan_2023_Learning}. In contrast to classical manually designed optimization techniques, learned optimizers are typically composed of black-box function approximators \cite{Metz_2020_Tasks,Vishnu_2020_MetaLearningBBO}, or they build on hand-designed update rules whose hyperparameters are learned~\cite{Kristiansen_2024_Narrowing,Moudgil_2025_Celo}. 
Despite promising results, an open challenge remains how to learn an optimizer's behavior that (i) reduces the need for extensive tuning across tasks and (ii) still offers adaptive learning rate schedules for quick convergence. 

\begin{figure*}[!ht]
    \centering
    \includegraphics[width=\textwidth]{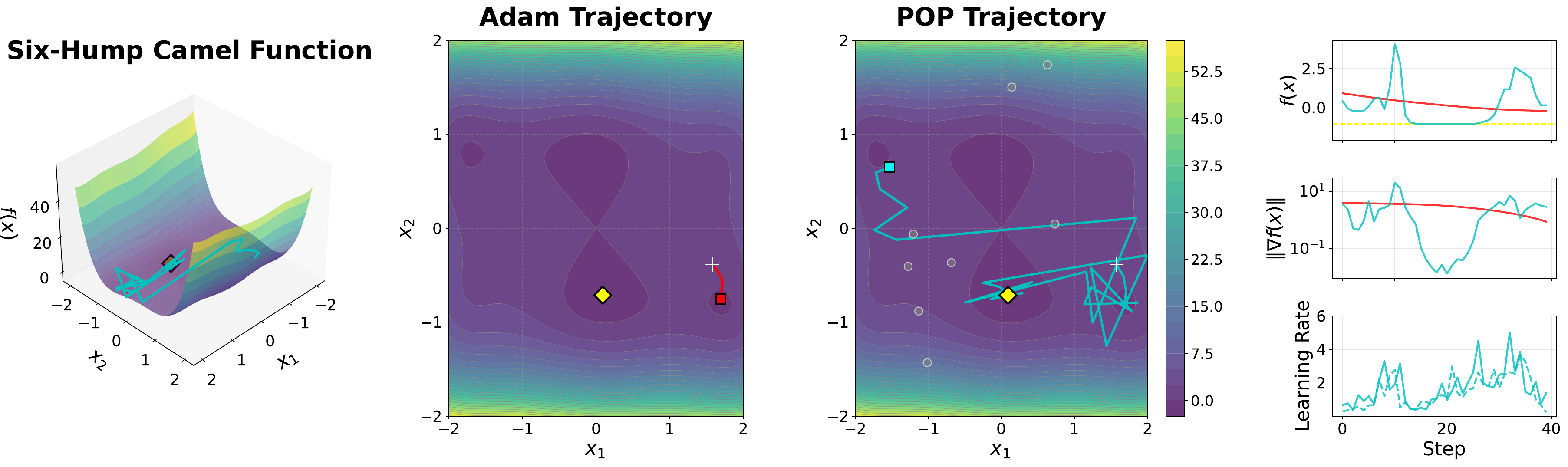}
    \caption{POP (blue) adapts its learning rate based on the optimization landscape, enabling rapid convergence, escape from local minima, and improved global optimization compared to Adam (red). Yellow diamond represents the global minima, the white cross represents the start state, and the square represents the end state.}
    \label{fig:motivation}
\end{figure*}

We introduce POP, a meta-learned optimizer trained as a continuous-control agent through Reinforcement Learning (RL) meta-learning with optimization-specific reward shaping. POP meta-learns a policy that outputs per-step learning rates conditioned on the optimization history, allowing it to adapt learning rates based on similarities between the current landscape and those encountered during meta-training. 
Our model leverages a transformer-based backbone \citep{Vaswani_2017_transformer} to encode long-range trajectory context. We meta-train our model on millions of diverse objective functions sampled from a novel prior over landscapes. Specifically, we construct this prior using a Gaussian process \citep{Rasmussen2006Gaussian}, approximated via Random Fourier Features \citep{Rahimi2007RandomFF}, that allows scalable training on a set of diverse objective landscapes at a low computational cost.

We illustrate the benefit of a context-conditioned optimizer in \Cref{fig:motivation}. POP increases the learning rate when the optimization landscape is sufficiently smooth, enabling rapid convergence to a local minimum. Upon reaching such a minimum, POP can either refine the solution through smaller updates or promote exploration by increasing the learning rate, leveraging contextual information gathered along the trajectory and patterns learned from similar functions during meta-training. 
This adaptive behavior emerges from the meta-learning phase, during which we train POP to optimize millions of diverse functions.

A thorough evaluation highlights POP's generalization on the \textit{Virtual Library of Simulation Experiments} \citep{simulationlib} benchmark, consisting of $43$ diverse optimization functions. We further evaluate our model's capabilities to transfer to longer optimization horizons and higher-dimensional problems. A statistical ranking analysis reveals that our method outperforms classical optimizers as well as a meta-learned optimizer introduced recently under matched budget constraints. The key contributions of this paper are:
\begin{itemize}
\item A novel optimizer that is meta-trained over millions of functions, learning step size policies conditioned on the optimization's contextual information, enabling effective exploration–exploitation behavior during optimization;
\item A novel RL reward formulation and function-scaling that ensures in-distribution generalization of the optimization policy;
\item A prior over optimization functions based on Gaussian processes that is used to sample millions of functions, thereby enabling scalable meta-training on diverse optimization landscapes;
\item An empirical evaluation on $43$ benchmark functions showing superior performance over conventional optimizers, with ablations on reward design, evaluation budget, and scalability to higher dimensions.
\end{itemize}

%% file: 03_methodology.tex
\section{POP: Prior-fitted Optimizer}
\label{ref:methodology}
\subsection{Problem Setting}
\label{subsec:problemSetting}
In the following, we consider unconstrained optimization problems of the form $x^\ast \in \text{argmin}_{x \in \mathbb{R}^d} f(x),$ where each task instance is an objective function $f$ sampled from a prior $f \sim p(f)$ over the space of compact functions. We refer to an episode as optimizing a single sampled function $f$ for a fixed budget of $T$ time steps. We define an optimization trajectory as:
\begin{definition}[Optimization trajectory]
\label{def:opt_traj}
For an objective function $f$, we define the  optimization trajectory that includes the first $t$ update steps of a maximum budget of $T$ updates as:
\begin{equation}
\tau^{(t)} \coloneqq \left\{\left(x_i, y_i, \nabla f(x_i), \frac{i}{T} \right)\right\}_{i=1}^t,    
\end{equation}
where $x_i$ denotes the parameters of the optimization problem in step $i$, while $y_i{\coloneqq}f(x_i)$ is the function value, $\nabla f(x_i)$ the gradient information, and $\frac{i}{T}$ the normalized timestep.
\end{definition}
First-order optimizers proceed by applying an update rule based on local gradient information, e.g., GD applies $x_{t}=x_{t-1} - \eta_{t}\nabla f(x_{t-1})$, where $\eta_{t}$ is the step size at iteration $t$.
In this work, we replace a preset step size schedule by a meta-learned policy $\eta_t\sim \pi_\theta(\eta_t | \tau^{(t-1)})$, where at each iteration, our optimizer outputs the next step size $\eta_t$ via a state-conditioned control signal $\tau^{(t-1)}$ based on the optimization trajectory (cf. \Cref{def:opt_traj}) up to step $t{-}1$. 

\subsection{MDP for Optimization}
\label{subsec:mdp}
We formulate the optimization problem as a Markov Decision Process (MDP) defined by the tuple $(\mathcal{S}, \mathcal{A}, \mathcal{P}, r, p_0, \gamma)$, where $\mathcal{S}$ are the states, $\mathcal{A}$ the actions, $\mathcal{P}$ the transition dynamics, $r: \mathcal{A} \times \mathcal{S} \rightarrow \mathbb{R}$ is the reward signal, $p_0$ is the probability distribution of initial elements in the optimization trajectory, and $\gamma$ denotes the discount factor. 
Our goal is to train a parameterized policy $\pi_\theta: \mathcal{S} \times \mathcal{A} \rightarrow \mathbb{R}^+$.

In our meta-learning setting, an episode corresponds to an optimization trajectory $\tau$ on a sampled task instance from a prior, i.e., $f \sim p(f)$. In the sequential decision problem, the policy receives a representation of the current optimization context and outputs a continuous coordinate-wise step size action $a_t =\eta_t\in \mathbb{R}^d >0$. 
%
The state $s_t \in \mathcal{S}$ is simply the trajectory of updates so far $s_t \coloneqq \tau^{(t)}, \text{ for } t \in [1,\dots, T]$.

\textbf{Optimization Step.~}
Given an optimization trajectory $\tau^{(t-1)}$, our stochastic policy is a network with parameters $\theta$ that outputs the update steps at iteration $t$ for each function dimension $\mu_\theta\left(\tau^{(t-1)}\right) \in \mathbb{R}^d$ and the standard deviation $\sigma_\theta \in \mathbb{R}^d$.
\begin{align}
\pi_\theta\left(\eta_t \; | \; \tau^{(t-1)}\right) = \mathcal{N}\left(\mu_\theta\left(\tau^{(t-1)}\right), \mathrm{diag}\left(\sigma_\theta^2\right)\right)
\end{align}

To conduct an optimization, POP applies three steps. First, it samples a coordinate-wise next step size from our policy network (Equation~\ref{eq:samplelr}). Then, POP conducts a gradient update (Equation~\ref{eq:updatestep}), and finally, it updates the trajectory (Equation~\ref{eq:updatetrajectory}).
\begin{align}
\label{eq:samplelr}
\eta_t &\sim \pi_\theta\left(\eta_t \; | \; \tau^{(t-1)}\right)  \\
\label{eq:updatestep}
x_{t} &\leftarrow x_{t-1} - \eta_t \nabla f(x_{t-1}) \\
\label{eq:updatetrajectory}
\tau^{(t)} &\leftarrow \tau^{(t-1)}\cup \left\{\left( x_{t}, y_{t}, \nabla f(x_{t}), \frac{t}{T}\right)\right\}
\end{align}
%
%
\textbf{Reward Signal.~}
We propose a dedicated reward function for our optimization, defined as the improvement to the best observed function value so far in a trajectory. Let $y^*_{t-1}{=}\min_{\left(\cdot, y, \cdot, \cdot \right) \in \tau^{(t-1)}}{y}$, we define:
\begin{align}
R\left(y_t, \tau^{(t-1)}\right) = \max\left(0, y^*_{t-1} - y_t\right).  
\end{align}
This reward assigns a positive signal only when an action yields a new observed minimum, directly linking policy optimization to objective improvement. Since non-improving steps receive no penalty, the agent can explore the function freely, and meaningful improvements guide learning.
\subsection{Meta-Learning Objective} We optimize the policy parameters $\theta$ to maximize the expected return over tasks sampled from a prior (cf. \Cref{subsec:prior}). We formulate the objective as follows:
\begin{align}
\max_{\theta}~\mathbb{E}_{f \sim p(f)}~\mathbb{E}_{\tau \sim \text{POP}(f; \theta)} \sum_{t=0}^{T}\gamma^t  R\left(y_t, \tau^{(t-1)}\right),
\end{align}
where $\tau \sim \text{POP}(f; \theta)$ is computed by running Equations~\ref{eq:samplelr}-\ref{eq:updatetrajectory} for $T$ steps on the function $f$ sampled from the prior. We train the policy parameters using the Proximal Policy Optimization (PPO) algorithm~\cite{schulman_2017_ppo}.

%% file: 04_prior.tex
\subsection{A Prior for Optimization Problems}
\label{subsec:prior}

To meta-train our method, we sample optimization problems from a prior distribution, $f \sim p(f)$. This distribution must be sufficiently diverse and complex to cover a wide range of optimization problems, 
enabling learned optimizers to generalize well across tasks. 
Therefore, we define a prior over optimization objectives using a combination of components designed to induce diversity in the generated objective landscapes. These components include a quadratic term, saddle- and monkey-saddle inducing terms, tilting and confinement operations, as well as an approximation to a Gaussian process prior with an RBF kernel via \textit{Random Fourier Features} (RFF) for computational efficiency \cite{Rahimi2007RandomFF}. 
Further details about the prior, including a visual illustration, are provided in \Cref{appx:app_b}.

\section{In-Distribution Generalization}


One of the major challenges when meta-learning an optimizer's policy is the mismatch of domains and ranges across functions, leading to out-of-distribution failures (i.e., naively meta-learned policies fail for functions having very similar shapes but different scales).  
To ensure that our policy will be meta-learned and transferred to matching functions under an in-distribution assumption, we transform the domains and ranges of all training and test functions identically. 

\textbf{Initial Context.~}POP trains parametric policies that output the next learning rate based on the optimization trajectory so far, transferring knowledge from previously-seen functions with a similar landscape. Therefore, it needs an initial trajectory as an input before it conducts the first update step. 

As the initial context for our policy at the beginning of the optimization, we sample $c \in \mathbb{N}^+$ observations.
Let $p_0 \in \mathcal{P}(x)$ denote a random initial probability measure on the parameter space, where $\mathcal{P}(x)$ is the set of probability distributions on a functions' parameter space. We
define $s_c \sim p_0 \Leftrightarrow\;s_c \coloneqq \tau^{(c)}$.
After sampling the initial random context, we use the $x_i$ with the lowest $f(x_i)$ as the starting point of the gradient descent update steps leveraging coordinate-wise step sizes.


\textbf{Boundary Scaling.~}
For each function, we assume known domain coordinate-wise boundaries $x \in [x^{\min}, x^{\max}]$ and map coordinates to a fixed symmetric interval via boundary scaling per dimension.
\begin{equation}
x_t^{\mathrm{bnd}} = V_x \cdot \left(2\cdot \frac{x_t - x^{\min}}{x^{\max} - x^{\min}} - 1\right) \quad \text{with~} V_x\in\mathbb{R}^+.
\end{equation}
where $V_x$ refers to the target scale.
Analogously, we scale function values using running extrema
$y_{t,\min} \coloneqq \min_{j \le t} y_j$ and
$y_{t,\max} \coloneqq \max_{j \le t} y_j$,
initialized from the random context $s_c$ and update them online when the function's landscape is explored and the context information increases: 
\begin{equation}
y_t^{\mathrm{bnd}} = V_y \cdot \left(2 \cdot \frac{y_t - y^{\min}_{t}}{y^{\max}_{t} - y^{\min}_{t}} - 1\right) \quad \text{with~} V_y\in\mathbb{R}^+,   
\end{equation}
where $V_y$ refers to the target scale.

\textbf{Z-transformation.~}Boundary-scaled $x_t^{\mathrm{bnd}}$ and $y_t^{\mathrm{bnd}}$ are then standardized by applying an online Z-transformation using the evaluation data of the current optimization trajectory, yielding $\tilde{x}_t$ and $\tilde{y}_t$.

\textbf{Gradient Scaling.~}Because $\tilde{x}_t,\tilde{y}_t$ are transformed before being passed to the model, gradients defined in the original coordinates are no longer consistent with the transformed space. We therefore apply the corresponding Jacobian to obtain gradients that reflect the correct local sensitivities:
\begin{equation}
\nabla \tilde{f}(x_t)
=
\left(\frac{\sigma_{x,t}}{\sigma_{y,t}}\right)
\left(\frac{V_y}{V_x}\right)
\left(\frac{x^{\max}-x^{\min}}{y^{\max}_{t}-y^{\min}_{t}}\right)
\odot \nabla f(x_t),
\end{equation}
with $V_x, V_y \in \mathbb{R^+}$ denoting coordinate-wise scaling parameters, and $\sigma_{x,t}, \sigma_{y,t}$ denoting the standard deviation of the respective dimension up to time step $t$. 
The optimizer maintains one state per coordinate, defined exclusively in the transformed space. 

\textbf{Transformed Update.~}
Since the model operates on the transformed variables $\tilde{\tau}^{(t-1)}$, the gradient update is performed in the transformed space. Specifically, $\tilde{x}_{t} = \tilde{x}_{t-1} - \eta_t \nabla \tilde{f}(x_{t-1})$. The updated iterate $\tilde{x}_{t}$ is then mapped back to the original domain by applying the inverse Z-transformation followed by inverse boundary scaling, after which it is evaluated using the original objective function $f(x_{t})$. Following the evaluation of the new $x_t$, $y_t$, and $\nabla f(x_{t})$, the entire transformed set $\tilde{\tau}^{(t)}$ is re-transformed using the updated statistics.

%% file: 05_experimental_protocol.tex
\section{Experimental Protocol}
\label{sec:exp_protocol}

We train our proposed optimizer (cf.~\Cref{ref:methodology}) on our newly proposed prior (cf.~\Cref{appx:app_b}), with parameter distributions detailed in~\Cref{05:prior_values} in Appendix \ref{appx:prior_params}. Notably, we restrict our prior to two-dimensional ($D{=}2$) objective functions. Visualizations of randomly sampled functions from this prior are shown in~\Cref{appx:prior_viz}. Rather than conditioning on the full optimization trajectory $\tau^{(t-1)}$, we decompose trajectories along dimensions and restrict the policy input to the history of the corresponding coordinate. The optimizer thus operates independently per dimension, enforcing conditional independence during training while enabling efficient batching and scalability to higher-dimensional problems.  We open-source our implementation~\footnote{\url{https://anonymous.4open.science/r/pop-F342/README.md}} to foster future research.

\subsection{Architecture and Training}
We parameterize both the actor and critic using a shared transformer backbone with separate MLP heads, for a total of $333K$ trainable parameters. Full architectural and training hyperparameters are reported in \Cref{tab:hyperparams,tab:ppo_hparams}. The actor outputs the mean of a Gaussian policy and samples actions using a learned, state-independent log standard deviation. For numerical stability, the log standard deviation is passed through a $\tanh$ nonlinearity and rescaled to lie in $[-3.0,\, 2.0]$. To enforce strictly positive learning rates, sampled actions are exponentiated via $\exp(\eta_t)$.

At each training iteration, we sample $B = 256$  optimization problems from the function prior. For each sampled function, we collect 16 rollout trajectories. Each rollout begins with a newly generated random initial context of $c=10$ points, after which the agent performs $T=40$ optimization steps. After all rollouts for the iteration are collected, the trajectories are combined and used to update the policy and value networks with PPO. We train POP for 8,000 iterations, yielding approximately $3.27 \times 10^{7}$ optimization problem instances. Training requires around 56 hours on a single NVIDIA H100 GPU.

\subsection{Benchmarks}
\textbf{In-distribution Performance.} We construct a validation set for ablations consisting of 1024 optimization problems in $D{=}2$, and test sets for one-shot evaluation with 1024 problems each for $D \in \{2, 16, 32\}$. All problems use a fixed initial context of size $c{=}10$.

\textbf{Out-of-distribution Experiments.} We further evaluate the out-of-distribution performance of our method on 43 optimization problems from the widely used Virtual Library of Simulation Experiments (VLSE) benchmark~\citep{simulationlib}. It covers a diverse range of optimization landscapes, including functions with many local minima, as well as bowl-shaped, plate-shaped, and valley-shaped geometries, in addition to steep ridges and drops.
Each benchmark problem is evaluated at its native dimensionality, which ranges from 1D to 6D across the suite. 

\subsection{Baselines}
We compare our method against a diverse set of gradient-based and learned optimization baselines. \textbf{Gradient Descent (GD)} \citep{cauchy1847},  \textbf{Adam} \citep{Kingma_2014_adam}, and \textbf{Lion} \citep{chen2023symbolic} represent first-order gradient-based optimizers, while \textbf{Sophia-G} \citep{liu2023sophia} is a second-order-inspired optimizer that uses a lightweight diagonal Hessian approximation. \textbf{L-BFGS} \citep{liu1989L-bfgs} is a quasi-Newton method leveraging approximate second-order information. We also add \textbf{Random Search (RS)} as a placebo method. Finally, \textbf{VeLo} is a meta-learned optimization technique from a prior over neural network training tasks. All baselines are evaluated under the same optimization budget.


%% file: 06_evaluation.tex
\section{Evaluation}
\label{sec:evaluation}

\paragraph{Hypothesis 1: Learned optimizers trained on synthetic priors can generalize to unseen problems from the same distribution.}\hfill\\
%
\begin{figure}[b]
\begin{minipage}[t]{0.5\textwidth}
    \includegraphics[width=0.9\linewidth, trim={0 0 0 1.0cm}, clip]{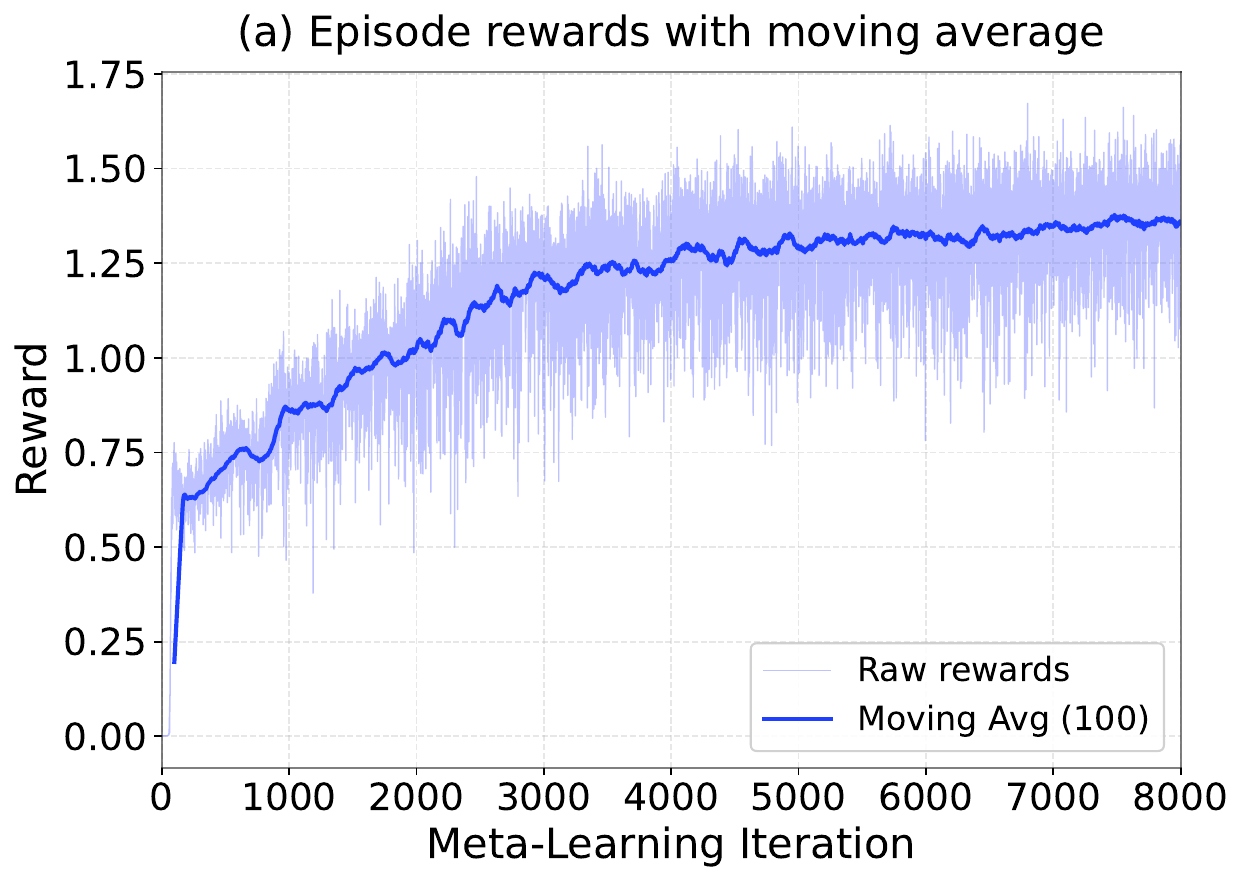}
    \captionof{figure}{Episode rewards with moving average}
    \label{fig:reward_eval_2A}
\end{minipage}
\begin{minipage}[t]{0.5\textwidth}
    \includegraphics[width=\linewidth]{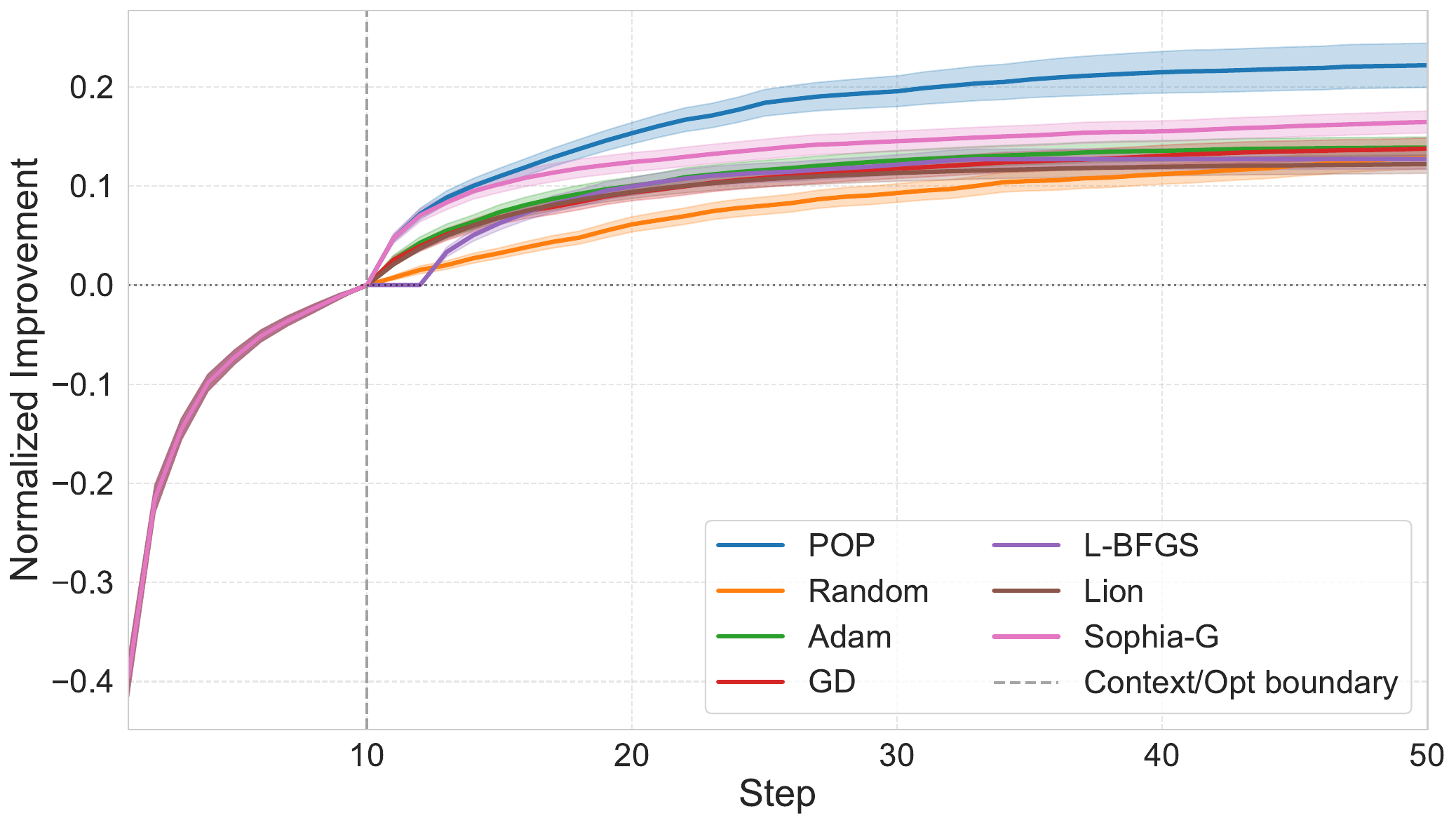}
    \captionof{figure}{In-distribution test set performance vs. baselines. Mean normalized improvement over steps; shading indicates 95\% CIs. Dashed line marks the context/optimization boundary.}
        \label{fig:in-distribution-comparison}
\end{minipage}
\end{figure}
We first examine the training dynamics of POP. Since each meta-RL episode is generated from a newly sampled objective function, the episode reward reflects the optimizer’s average performance over the task distribution rather than repeated optimization on a fixed set of functions. As shown in~\Cref{fig:reward_eval_2A}, the reward increases during training and eventually stabilizes, indicating that POP converges to a stable optimization policy. We further observe that POP adapts its behavior to the local structure of the objective landscape. Rather than following a fixed search pattern, it alternates between exploratory moves across the domain and localized refinement near promising regions. The resulting trajectories show that POP can escape suboptimal areas, revisit promising basins, and adjust its step sizes over time, indicating that its optimization strategy is conditioned on the context. A visual illustration of this behavior is provided in \Cref{appd:in_dist}.

Next, we evaluate the competitiveness of our method by comparing it against a set of gradient-based optimization baselines, including GD, Adam, L-BFGS, Lion, Sophia-G, and random search, on unseen test problems drawn from the same prior distribution. Since we require $c$ initial context points, we initialize all methods with identical context sets. For the optimization baselines, we select the best-performing context point as the initialization. We tune the learning rate for GD, Adam, L-BFGS, Lion, and Sophia-G using grid search on the validation set. Hyperparameter settings are detailed in~\Cref{app_c}. We evaluate all methods on the test set.
We measure performance by normalized improvement (NI):
\[
\mathrm{NI}
=
\frac{y^{\min}_{t} - y^*_{t}}
{y^{\max}_{t} - y^{\min}_{t} + \epsilon},
\]
where \(y^{\min}_{t}\) and \(y^{\max}_{t}\) are the minimum and maximum losses in the initial context, \(y^*_{t}\) is the best loss found so far, and \(\epsilon\) ensures numerical stability. 
%
%

%
%

\begin{figure}
\centering
\begin{minipage}[b]{0.495\textwidth}
  \includegraphics[width=\linewidth]{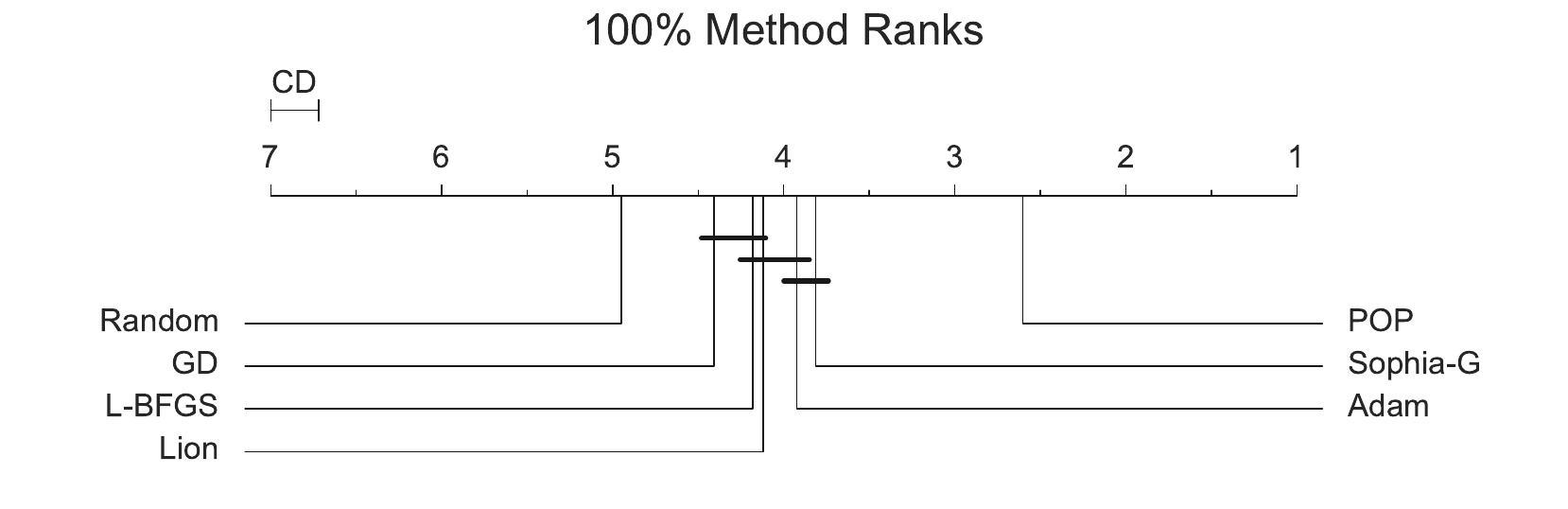}
  \captionof{figure}{Rankings on the in-distribution test set at 100\% budget. Lower ranks correspond to better performance, while horizontal bars indicate differences that are not statistically significant.}
  \label{fig:in-distribution-rank}
\end{minipage}
\hfill
\begin{minipage}[b]{.495\textwidth}
  \centering
  \includegraphics[width=\columnwidth]{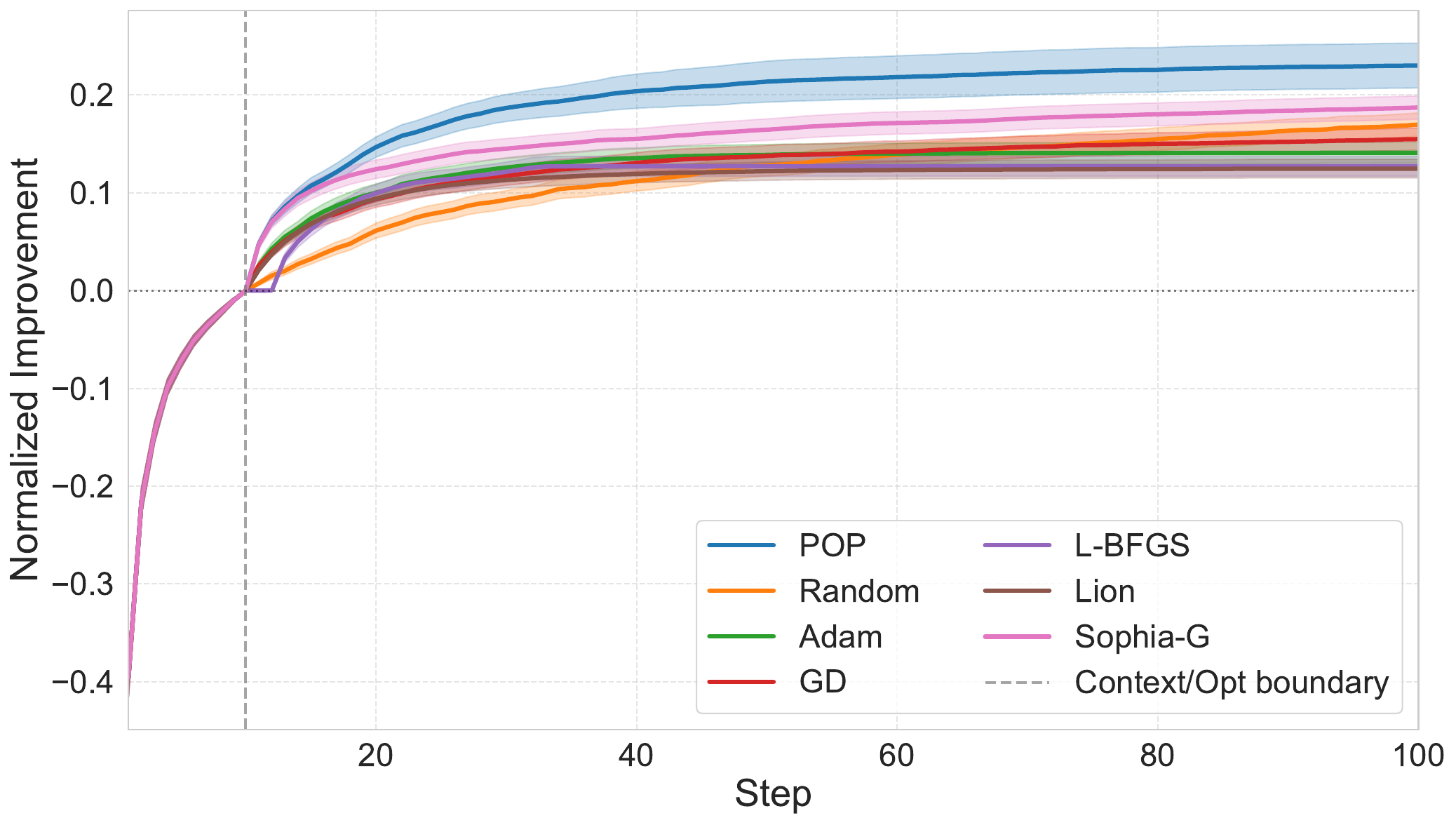}
  \captionof{figure}{In-distribution test set performance vs. baselines at twice the training budget. Mean normalized improvement over steps; shading indicates 95\% CIs. Dashed line marks the context/optimization boundary.}
  \label{fig:2D_100}
\end{minipage}
\end{figure}

As shown in~\Cref{fig:in-distribution-comparison,fig:in-distribution-rank}, POP outperforms all baselines by achieving higher normalized improvement during optimization. Gradient-based baselines, including GD, Adam, L-BFGS, Lion, and Sophia-G, tend to plateau earlier, while random search improves more slowly. It is worth mentioning that POP's novel RL pretraining leads to an emerging ability to escape local optima, which is not observed in prior gradient-based methods. We provide a comparison against non-convex solvers in~\Cref{app_h:h1}, demonstrating that POP is even competitive against non-convex solvers on multi-optima functions, despite being a first-order gradient approach. 



\textbf{Ablating the Reward Formulation.~} We compare five reward functions that differ in how they incentivize improvement using a lightweight model variant with 166K parameters. Let $\tilde{y}_t$ denote the boundary scaled and Z-transformed current objective value and $\tilde{y}^*_{t-1}$ the boundary scaled and Z-transformed best value observed in the optimization trajectory. The corresponding reward functions are defined as~:
%
%
\begin{equation*}
\small
\begin{array}{lll}
R_{\text{Current}} = \tilde{y}_t,
&
R_{\text{Global Imp}} = \tilde{y}^*_{t-1} - \tilde{y}_t,
&
R_{\text{Global Imp}_c} =
\text{max}\bigl(0, \tilde{y}^*_{t-1} - \tilde{y}_t\bigr),
\\[0.4em]
R_{\text{SMAPE}} =
\text{max}\bigl(0, 
\frac{2(\tilde{y}^*_{t-1} - \tilde{y}_t)}
     {|\tilde{y}^*_{t-1}| + |\tilde{y}_t|}
\bigr),
&
\multicolumn{2}{l}{
R_{\text{Mix},\alpha} =
\alpha\text{max}\bigl(0, \tilde{y}^*_{t-1} - \tilde{y}_t\bigr)
+ (1-\alpha)(\tilde{y}_{t-1}-\tilde{y}_t).
}
\end{array}
\end{equation*}

\begin{figure}[t]
\begin{minipage}[c]{0.4\textwidth}
\centering
\captionof{table}{Reward ablation.}
\begin{adjustbox}{max width=0.9\textwidth}
\begin{tabular}{l S[table-format=1.4] S[table-format=1.4] S[table-format=1.4]}
\toprule
Method & {Mean} & {CI$_{95\%}$ low} & {CI$_{95\%}$ high} \\
\midrule
$R_{\text{Current}}$          & 0.1522 & 0.1403 & 0.1641 \\
$R_{\text{Global Imp}}$       & 0.1120 & 0.1038 & 0.1202 \\
$R_{\text{Global Imp}_{c}}$   & {\bfseries 0.1990} & 0.1843 & 0.2138 \\
$R_{\text{SMAPE}}$            & 0.1669 & 0.1562 & 0.1775 \\
$R_{\text{Mix},\alpha=0.1}$   & 0.1354 & 0.1241 & 0.1467 \\
$R_{\text{Mix},\alpha=0.2}$   & 0.1385 & 0.1268 & 0.1503 \\
$R_{\text{Mix},\alpha=0.3}$   & 0.1401 & 0.1283 & 0.1519 \\
\bottomrule
\end{tabular}
\end{adjustbox}
\label{tab:Reward_ablation}
\end{minipage}
\begin{minipage}{.575\textwidth}
  \centering
  {\includegraphics[width=\columnwidth]{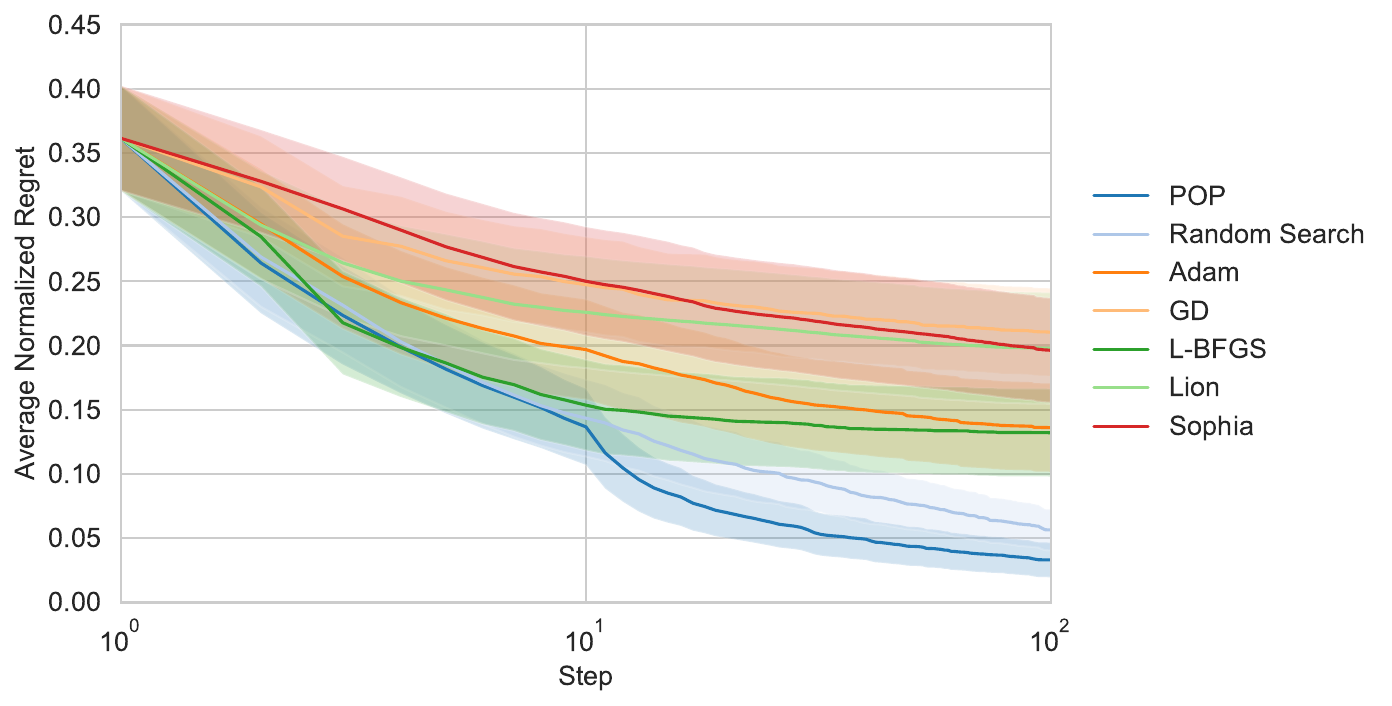}
  \captionof{figure}{The average normalized regret for all methods on VLSE. Solid lines represent the mean value.}
  \label{fig:hyp_3_average_normalized_regret}
  }
\end{minipage}
\end{figure}

We train each model for 2K iterations and evaluate each on the within-distribution validation set with identical starting context.~\Cref{tab:Reward_ablation} shows that $R_{\text{Global Imp}_c}$ achieves the highest mean normalized improvement. We attribute this to the fact that clipped rewards do not penalize exploratory actions, allowing the agent to temporarily move away from the current best solution without incurring negative reward, enabling to find better solutions.

\paragraph{Hypothesis 2: Learned coordinate optimizers trained on a low-dimensional (2D) prior for a fixed number of iterations generalize to longer horizons and higher-dimensional problems.}\hfill\\
To test whether the POP optimizer trained on 2D trajectories generalizes beyond its training horizon, we evaluate performance under a longer optimization budget on the test functions. Since our architecture uses last-token pooling, the model is not restricted to the sequence length observed during training. We therefore extend the evaluation budget from 50 to 100 iterations, setting c=10 and allowing the model to predict the subsequent T=90 updates.

As shown in~\Cref{fig:2D_100}, POP continues to improve beyond the training horizon, demonstrating strong extrapolation to longer optimization trajectories. Under the 100-step budget, POP attains the best overall performance, outperforming all gradient-based baselines, including GD, Adam, L-BFGS, Lion, and Sophia-G. This indicates that POP’s learned optimization strategy generalizes effectively beyond the trajectory lengths observed during training.

Next, we evaluate whether POP generalizes to higher-dimensional test problems than the 2D problems encountered during training. Due to its coordinate-wise formulation, POP can be applied to higher-dimensional problems by batching across coordinates, without architectural modification or retraining. We evaluate the model on 16D and 32D test problems, running each method for 50 optimization steps\footnote{To maintain comparable smoothness across dimensions, the number of parameters is increased to $M \in \{16000, 32000\}$.}. For all baselines, we zero-shot apply the best learning rates found on the 2D validation set. Additional results for an extended set of optimization methods, covering both the 100-iteration setting and the higher-dimensional experiments, are provided in~\Cref{app_h:h2}.



\begin{table}[t]
\centering
\caption{Mean performance across different dimensionalities.}
\resizebox{0.7\columnwidth}{!}{
\begin{tabular}{lccccccc}
\hline
Dim. & POP & Random & Adam & GD & L-BFGS & Lion & Sophia-G \\
\hline
16D & \textbf{0.2911} & 0.1177  &  \underline{0.2887} & 0.2510 & 0.1728 & 0.2409 & 0.2579 \\
32D & \underline{0.4966}  & 0.1669 & \textbf{0.5029} & 0.4003 & 0.4171 & 0.4245 & 0.4820\\
\hline
\end{tabular}
}
\label{tab:dimensionality_results}
\end{table}%

%
%
 
\paragraph{Hypothesis 3: Learned coordinate optimizers trained on a low-dimensional (2D) synthetic prior can generalize to out-of-distribution problems.}\hfill\\
We further validate the generalization capabilities of our method on the \textit{Virtual Library of Simulation Experiments}~\citep{simulationlib} benchmark.
Since the optimal hyperparameters for each algorithm on the benchmark are unknown, all baseline methods are evaluated using their default settings. Initially, we compare all methods in terms of per-task normalized regret $r_{\text{opt}, t}
= \frac{y^*_{t} - y^{\text{min}}}
{y^{\text{max}} - y^{\text{min}}}$, where $y^*_{t}$ represents the best function value achieved by an optimizer at iteration $t$ for a given function, while $y^{\text{min}}$ ($y^{\text{max}}$) represent the minimal (maximal) value of that function.

In ~\Cref{fig:hyp_3_average_normalized_regret}, we present the average normalized regret for all methods across the different optimization tasks, where POP manages to significantly outperform all the baselines.
%
%
%

\begin{figure}
    \centering
    \begin{subfigure}{0.495\textwidth}
        \centering
        \includegraphics[width=\textwidth, trim={0 0 0 1cm}, clip]{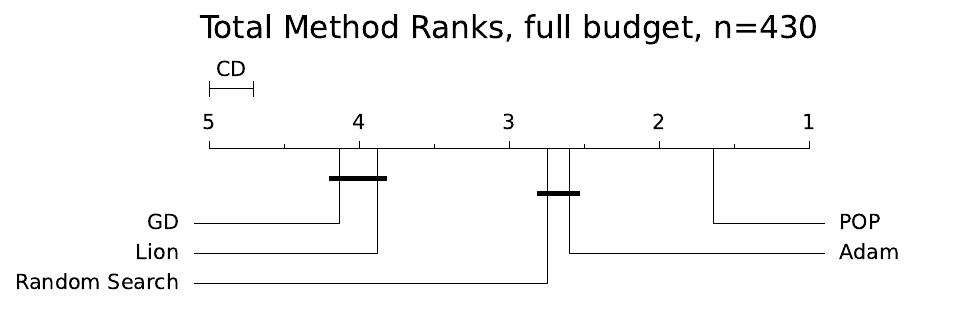}
        \caption{POP vs. First-order Methods}
        \label{subfig:pop_vs_first}
    \end{subfigure}
    \begin{subfigure}{0.495\textwidth}
        \centering
        \includegraphics[width=\textwidth, trim={0 0 0 1cm}, clip]{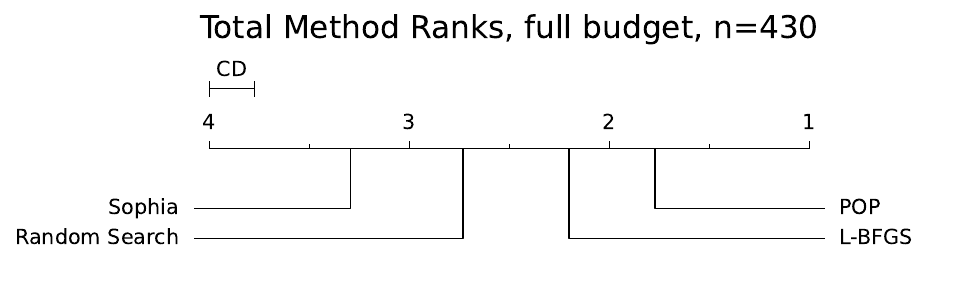}
        \caption{POP vs. Second-order Methods}
        \label{subfig:pop_vs_second}
    \end{subfigure}
    \caption{Performance of POP compared across different model categories and representative models.}
    \label{fig:ood_categories}
\end{figure}

Furthermore, in \Cref{fig:ood_categories} we present critical difference (CD) diagrams to investigate method rankings and the statistical significance of the results. To build the CD diagrams, we use the autorank~\citep{Herbold2020} package that runs a Friedman test followed by a Nemenyi post-hoc test at a 0.05 significance level.
\Cref{fig:ood_categories} shows a more detailed comparison of POP across different model families and their representative methods. Notably, POP shows superior performance and achieves the lowest rank among the first-order and second-order methods. A comparison to to further model families can be found in \Cref{app_h}.




%% file: 02_related_work.tex
\section{Related Work}
\label{sec:related_works}

\textbf{Optimizers.}
Optimization aims to find the minimum/maximum of a given objective function. Depending on the underlying structure of the function, different optimization methods may be more or less effective. For convex functions, second-order methods such as Newton’s method \cite{newton1736method} or quasi-Newton approaches like L-BFGS \cite{liu1989L-bfgs} can be highly efficient. In contrast, for non-convex problems, population-based methods such as \textit{Genetic Algorithms} \citep{holland1975adaptation} and \textit{Differential Evolution} \citep{storn1997differential} are often effective at escaping local minima. For low-dimensional problems, \textit{Bayesian Optimization} \cite{bergstra2011algorithms} can be particularly powerful, although it scales poorly with increasing dimensionality. In higher-dimensional settings, such as training neural networks, first-order methods including \textit{Gradient Descent} \cite{cauchy1847} and adaptive variants like \textit{Adam} \cite{Kingma_2014_adam} are widely used.

\textbf{Learned Optimizers.}
A growing field that tries to automate the design and tuning of optimization algorithms is known as \textit{Learning to Optimize} (L2O) \cite{Chen_2021_Learning, Tang_2024_L20Overview}. Most L2O approaches meta-train optimizers on NN training tasks \cite{Metz_2022_VeLO,xu2019learning, Meta-LR-Schedule-Net}. These methods differ in what aspects of the optimization process are learned, ranging from direct parameter updates \cite{Metz_2022_VeLO} to the prediction of a global learning rate schedule \cite{xu2019learning, Meta-LR-Schedule-Net}. They also vary in their meta-learning methodology, with some approaches relying on reinforcement learning \cite{xu2019learning}, while others employ gradient-based meta-training \cite{Metz_2022_VeLO,gärtner2023transformerbasedlearnedoptimization}.

\textbf{Prior-fitted Models.} 
Prior-Data Fitted Networks (PFN) \cite{mueller_2022_pfn, Hollmann_2025_Accurate} was introduced as a transformer-based model that is meta-learned by training on millions of synthetic datasets generated from specified priors.

\textbf{Novelty and Delineation from Prior Work}\\
Our approach, POP, differs from prior work along several key dimensions. Instead of meta-training on neural network losses \cite{Meta-LR-Schedule-Net, xu2019learning, Metz_2022_VeLO}, we train on a novel prior over convex and non-convex optimization problems, enabling meta-training over millions of synthetic functions at low computational cost. Rather than learning a global learning-rate schedule, POP predicts coordinate-wise learning rates. Using carefully designed input transformations such as boundary scaling, Z-transformation, and gradient scaling, POP exhibits strong generalization. Finally, coordinate-wise independence allows POP to scale naturally to high-dimensional settings via batching, while the transformer backbone efficiently processes long per-coordinate optimization histories.

Our work falls within meta-learning optimizers, where a model is trained to solve a distribution of optimization problems and generalize to unseen ones. We employ a transformer-based backbone that processes optimization trajectories and gradients to predict adaptive step sizes, effectively replacing classical solvers such as gradient descent. Operating at the algorithmic level, the learned model acts as a controller over the optimization process, modulating update directions and removing the need for hand-designed learning rate schedules.

%% file: 07_conclusion.tex
\section{Limitations and Future Work}
\label{sec:limitation}
\textbf{Limitations.} In this paper, we showcased the potential that meta-learned RL has in offering a strongly performing adaptive learning rate mechanism. However, our method is still a proof-of-concept and requires further research to mature into a scalable all-purpose optimizer. First of all, in our framework, we rely on a trajectory-conditioned decision that requires contextual information from the optimization history. Following the definition from \Cref{subsec:problemSetting}, POP requires positional information, function values, gradient information, and normalized timesteps. This results in a linear growth in the memory footprint with the number of contextual information being provided to POP. In our proof of concept, we rely on rather short horizons, forcing the optimizer to be efficient in its exploration of minima. However, increasing the optimization trajectory comes along with retaining also per-coordinate histories and inducing higher computational costs in the transformer attention cost over the stored sequence. Therefore, the focus in this work is on showing the capability of RL-enhanced optimization strategies rather than optimizing the computational and memory costs. Currently, the definition of our prior is also tailored towards low-dimensional continuous function optimization, and therefore, is also not yet applicable to solve diverse ML downstream tasks, where the objective landscapes substantially differ in stochasticity, curvature, scale, and non-stationarity. 

\textbf{Future Work.~} While our proposed method works well for optimization, its application to machine learning tasks remains an open direction. Therefore, our primary goal for the future is to extend the definition of our prior and make POP applicable also to ML tasks. A key challenge is designing synthetic priors that better reflect the structure of ML loss landscapes and generalize beyond function optimization. In addition, we plan to train the method on longer optimization trajectories and develop more compute-efficient policy architectures.


\section{Conclusion}
\label{sec:conclusion}
Optimization algorithms are central to minimizing functions, yet gradient-based methods that rely on first-order information often become trapped in local minima, limiting their ability to explore complex landscapes.
We propose POP, a meta-learned optimizer that learns step size policies conditioned on the accumulated optimization history, enabling strategic escapes from local minima. POP is meta-trained on millions of synthetic optimization problems drawn from a novel Gaussian process prior, covering both convex and non-convex regimes. In a thorough evaluation, our model generalizes well and achieves strong performance on a well-established benchmark suite encompassing 43 diverse optimization functions. Overall, POP shows that state-conditioned learning rate policies, meta-trained on synthetic priors, enable robust and general-purpose optimization capabilities.

%% file: 08_gp_proof.tex
\section{On the Universality of Gaussian Process Prior for
Pretraining Learnable Optimizers}
\label{app:proof}
The statement that a Gaussian Process (GP) is a universal approximator is a consequence of the properties of its covariance function, or kernel. The universality is not inherent to all GPs but depends on the choice of a \textit{universal kernel}. The proof relies on the connection between GPs and Reproducing Kernel Hilbert Spaces (RKHS). The approximation of the GP with Random Fourier Features introduces an additional approximation term absorbed into $\epsilon$.

\textbf{Theorem.}
Let $K \subset \mathbb{R}^d$ be a compact set. Let $k : K \times K \to \mathbb{R}$ be a universal kernel.
Let $f \in C(K)$ be any continuous function on $K$. Then, for any $\epsilon > 0$, there exists a set of points $X_n = \{x_1,\ldots,x_n\} \subset K$ and corresponding values $Y_n = \{f(x_1),\ldots,f(x_n)\}$ such that the posterior mean $\mu_n(x)$ of a Gaussian Process with prior mean $m(x)=0$ and kernel $k(x,x')$ satisfies:
$$
\sup_{x\in K} \bigl|f(x) - \mu_n(x)\bigr| < \epsilon.
$$

\subsection{Preliminaries: Gaussian Processes and RKHS}
A Gaussian Process $\mathrm{GP}(m(x), k(x,x'))$ is a stochastic process where any finite collection of random variables has a multivariate normal distribution. It is fully specified by its mean function $m(x)$ and covariance function (kernel) $k(x,x')$. For simplicity, we assume a zero prior mean, $m(x)=0$.

Given a set of $n$ noiseless observations $(X_n, Y_n)=\{(x_i, y_i)\}_{i=1}^n$ where $y_i = f(x_i)$, the posterior distribution of the GP at a new point $x^\ast$ is also Gaussian. The posterior mean is given by:
$$
\mu_n(x^\ast) \;=\; k(x^\ast, X_n)\,K_n^{-1}\,Y_n,
$$
where $k(x^\ast, X_n) = [k(x^\ast, x_1),\ldots,k(x^\ast, x_n)]$ is a row vector and $K_n$ is the $n\times n$ Gram matrix with entries $(K_n)_{ij} = k(x_i, x_j)$.

Associated with every positive definite kernel $k$ is a unique Reproducing Kernel Hilbert
Space (RKHS), denoted $\mathcal{H}_k$. This is a Hilbert space of functions on $K$ with two
key properties:
\begin{enumerate}
  \item For every $x\in K$, the function $k_x(\cdot) := k(x,\cdot)$ is in $\mathcal{H}_k$.
  \item (\emph{Reproducing property}) For any $g\in \mathcal{H}_k$ and $x\in K$, we have
  $\langle g, k_x\rangle_{\mathcal{H}_k} = g(x)$.
\end{enumerate}

A crucial result from GP theory is that the posterior mean $\mu_n(x)$ is the function in
$\mathcal{H}_k$ that minimizes the RKHS norm $\|g\|_{\mathcal{H}_k}$ subject to the
interpolation constraints $g(x_i)=y_i$ for all $i=1,\ldots,n$.

\subsection{Universal Kernels}
The approximation power of a GP is entirely determined by the richness of its RKHS. This
leads to the definition of a universal kernel.

\textbf{Definition (Universal Kernel).}
A continuous kernel $k$ on a compact set $K$ is called \emph{universal} if its corresponding
RKHS $\mathcal{H}_k$ is dense in the space of continuous functions $C(K)$ with respect to the
uniform norm ($L^\infty$).

This means that for any continuous function $f\in C(K)$ and any $\delta>0$, there exists a
function $g\in \mathcal{H}_k$ such that:
$$
\sup_{x\in K} |f(x) - g(x)| < \delta.
$$

Examples of universal kernels include:
\begin{itemize}
  \item \textbf{Gaussian (RBF) Kernel:}
  $$
  k(x,x') = \exp\!\left(-\frac{\|x-x'\|^2}{2\sigma^2}\right).
  $$
  \item \textbf{Mat\'ern Kernels:}
  $$
  k(x,x') = \frac{2^{1-\nu}}{\Gamma(\nu)}
  \left(\frac{\sqrt{2\nu}\,\|x-x'\|}{\ell}\right)^\nu
  K_\nu\!\left(\frac{\sqrt{2\nu}\,\|x-x'\|}{\ell}\right), \qquad \nu>0,
  $$
  where $K_\nu$ is the modified Bessel function of the second kind.
  \item \textbf{Laplacian Kernel:}
  $$
  k(x,x') = \exp\!\left(-\frac{\|x-x'\|}{\sigma}\right).
  $$
\end{itemize}

The universality of these kernels is often established using characterizations like
Bochner's theorem (a kernel is universal if its Fourier transform is positive on the entire
frequency domain).

\subsection{Proof of the Theorem}
Let $k$ be a universal kernel on a compact set $K$, and let $f\in C(K)$ be the target
function. Let $\epsilon>0$ be the desired precision.

\subsection*{Step 1: Density of RKHS in $C(K)$}
Since $k$ is a universal kernel, its RKHS $\mathcal{H}_k$ is dense in $C(K)$. Therefore, for
our given $\epsilon>0$, there exists a function $g\in \mathcal{H}_k$ such that:
$$
\sup_{x\in K} |f(x) - g(x)| < \frac{\epsilon}{2}.
$$

\subsection*{Step 2: Convergence of the Interpolant}
Now we need to show that we can choose data points $X_n$ such that the GP posterior mean
$\mu_n(x)$ (which interpolates $f$ on $X_n$) becomes arbitrarily close to $g(x)$.

Let $g$ be the function from Step 1. Since $g\in \mathcal{H}_k$, it has a finite RKHS norm,
$\|g\|_{\mathcal{H}_k} < \infty$. Let us choose a sequence of points
$X_n = \{x_1,\ldots,x_n\}$ that becomes dense in $K$ as $n\to\infty$.

Let $\mu_n^{g}$ be the GP posterior mean interpolating the values of $g$ on $X_n$, i.e.,
$\mu_n^{g}(x_i)=g(x_i)$. It is a standard result in the theory of RKHS (see Wendland,
\emph{Scattered Data Approximation}) that if $X_n$ becomes dense in $K$, then the sequence of
interpolants $\mu_n^{g}$ converges uniformly to $g$:
$$
\lim_{n\to\infty} \sup_{x\in K} |\mu_n^{g}(x) - g(x)| = 0.
$$
Therefore, we can choose $n$ large enough such that:
$$
\sup_{x\in K} |\mu_n^{g}(x) - g(x)| < \frac{\epsilon}{2}.
$$

\subsection*{Step 3: Combining the Approximations}
Let $\mu_n^{f}$ be the GP posterior mean interpolating the target function $f$ on the same
set of points $X_n$, i.e., $\mu_n^{f}(x_i)=f(x_i)$.

The posterior mean is a linear operator on the data values. Let
$$
L_n(Y)(x) = k(x, X_n)K_n^{-1}Y.
$$
Then $\mu_n^{f} = L_n(f(X_n))$ and $\mu_n^{g} = L_n(g(X_n))$.

We have:
\[
\sup_{x\in K} |\mu_n^{f}(x) - \mu_n^{g}(x)|
= \sup_{x\in K} |L_n(f(X_n) - g(X_n))(x)|
\le \|L_n\|_\infty \max_{i=1,\ldots,n} |f(x_i) - g(x_i)|.
\]

While $\|L_n\|_\infty$ can grow with $n$, we can proceed using the triangle inequality:
$$
\sup_{x\in K} |f(x) - \mu_n^{f}(x)|
\le
\sup_{x\in K} |f(x) - g(x)|
+
\sup_{x\in K} |g(x) - \mu_n^{g}(x)|
+
\sup_{x\in K} |\mu_n^{g}(x) - \mu_n^{f}(x)|.
$$
We have already bounded the first two terms. For the third term:
$$
\mu_n^{g}(x) - \mu_n^{f}(x) = \sum_{i=1}^n c_i\,k(x, x_i),
$$
where the coefficients satisfy
$$
c = K_n^{-1}\bigl(g(X_n) - f(X_n)\bigr).
$$
Hence,
$$
\sup_{x\in K} |\mu_n^{g}(x) - \mu_n^{f}(x)|
\le C_n \max_i |g(x_i) - f(x_i)|
< C_n\,\frac{\epsilon}{2},
$$
where $C_n$ depends on $X_n$.

A more elegant argument considers the projection of $f$ onto the subspace spanned by $\{k(x_i,\cdot)\}_{i=1}^n$. The posterior mean $\mu_n^{f}$ is precisely this projection. As $n\to\infty$ and $X_n$ becomes dense in $K$, the span of $\{k(x_i,\cdot)\}_{i=1}^n$ becomes dense in $\mathcal{H}_k$.

Thus, the sequence $\mu_n^{f}$ converges to the projection of $f$ onto $\mathcal{H}_k$, denoted $P_{\mathcal{H}_k}f$. So, for large enough $n$:
\[
\sup_{x\in K} |\mu_n^{f}(x) - P_{\mathcal{H}_k}f(x)| < \frac{\epsilon}{2}.
\]
By the property of projections, $P_{\mathcal{H}_k}f$ is the best approximation to $f$ within $\mathcal{H}_k$. Since $\mathcal{H}_k$ is dense in $C(K)$, we can find $g\in\mathcal{H}_k$ such that $\|f-g\|_\infty < \epsilon/2$. This implies that $\|f - P_{\mathcal{H}_k}f\|_\infty < \epsilon/2$. Combining these gives:
$$
\sup_{x\in K} |f(x) - \mu_n^{f}(x)|
<
\frac{\epsilon}{2} + \frac{\epsilon}{2}
=
\epsilon.
$$

\subsection*{Conclusion}
The proof establishes that for any continuous function $f$ and any precision $\epsilon$, one can find a finite set of sample points from $f$ such that the posterior mean of a GP with a universal kernel approximates $f$ to within $\epsilon$ over the entire compact domain. The universality of the GP is therefore inherited directly from the denseness of its associated RKHS in the space of continuous functions. This is a powerful existence result, though it does not specify the number of points $n$ required or the computational cost.
\newpage

%% file: 09_appendix_b.tex
\section{Synthetic Prior}
\label{appx:app_b}
\begin{figure}[h]
    \centering
    \includegraphics[width=14cm,height=21cm]{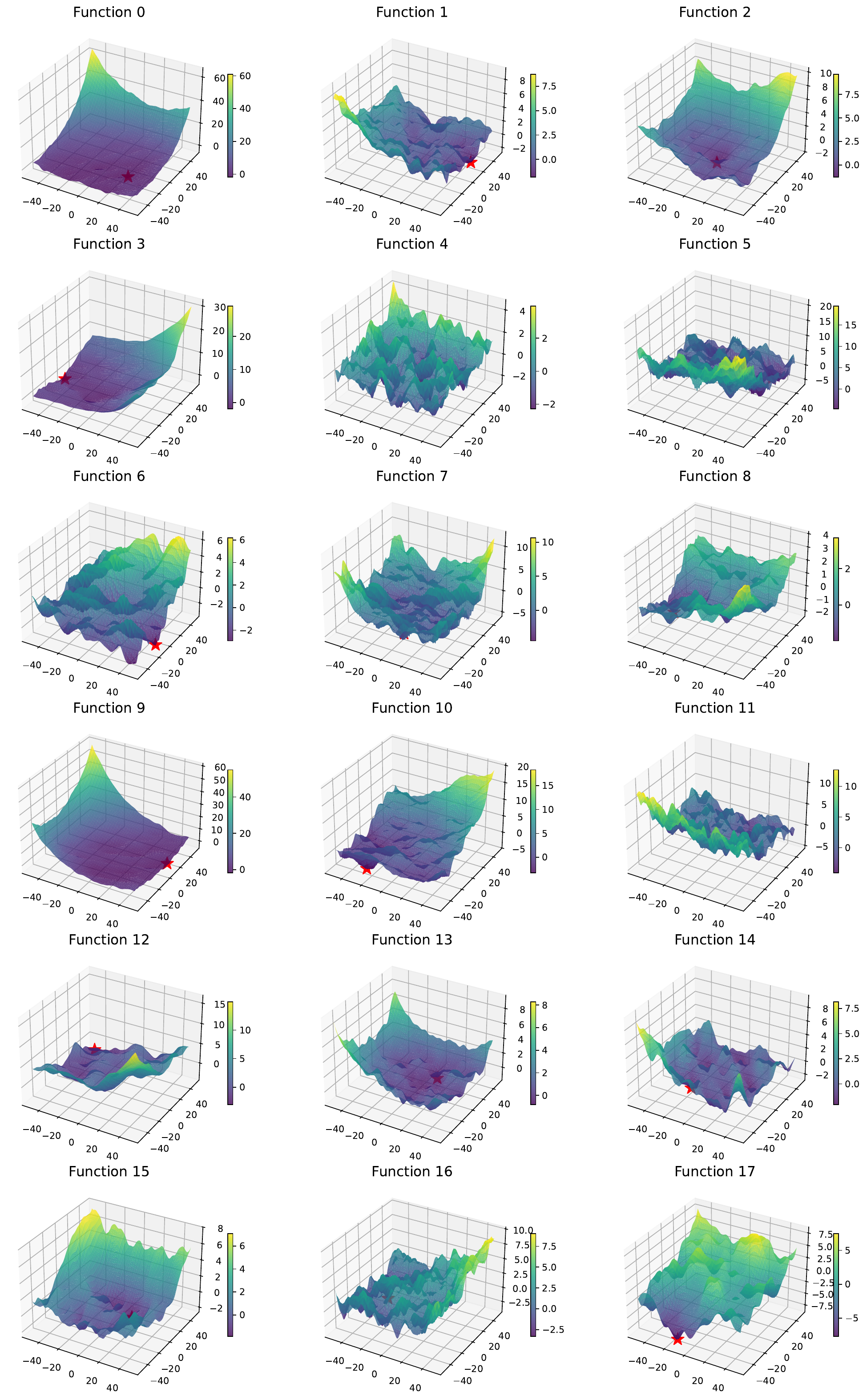}
    \caption{Exemplary functions sampled from the prior distribution (cf. \Cref{subsec:prior})}
    \label{fig:prior}
\end{figure}

This appendix details the synthetic prior over optimization landscapes used
for meta-training. We describe the functional form and its components,
summarize the sampled parameter ranges (cf. \Cref{05:prior_values}), and
provide random visual examples of drawn objectives (cf. \Cref{fig:prior}).

\subsection{Prior for Optimization Problems}\label{sec:prior_description}

We build each function from seven additive components. For a given number of dimensions~$D$, the function takes the form:
\begin{equation*}
\begin{aligned}
f(x) ={}&
\alpha_1 f_{\mathrm{RFF}}^{(1)}(x)
+ \alpha_2 f_{\mathrm{RFF}}^{(2)}(x)
+ \alpha_3 f_Q(x)
+ \alpha_4 f_S(x)
+ \alpha_5 f_M(x) \\
&+ f_{\mathrm{conf}}(x)
+ f_{\mathrm{tilt}}(x),
\end{aligned}
\end{equation*}
The weights
$(\alpha_1,\dots,\alpha_5)$ are initialized by drawing five independent positive
uniform random variables and normalizing them to sum to one. We then adjust these
weights according to the sampled landscape type: bowl-dominated instances suppress
the saddle and monkey-saddle terms, saddle instances receive a non-negligible saddle
weight, and instances without a monkey saddle set the corresponding weight to zero.
The remaining active weights are re-normalized where necessary. The confinement, tilt,
and vertical offset terms are added separately. 
In the following, we define each component in detail:

\paragraph{Quadratic component.}
This term adds a convex bowl-shaped surface:
\begin{equation*}
f_Q(x)
=
\sum_{d=1}^{D}
\beta^1_d
\bigl(R_Q \tilde{x}\bigr)_d^2 ,
\end{equation*}
where $\tilde{x}$ denotes the shifted coordinates normalized by the domain half-width,
$\beta^1_d$ sets the curvature along each dimension, and~$R_Q$ is an optional rotation
matrix. $\beta^1_d$ sets the curvature along each dimension and may be sampled with a target condition number to produce ill-conditioned objectives.

\paragraph{Saddle component.}
This term acts on independently rotated coordinates $(u,v)=R_S\tilde{x}$:
\begin{equation*}
f_S(x)
=
\tfrac{1}{2}\,\gamma^1
\bigl(u^2 - \gamma^2 v^2\bigr),
\end{equation*}
where $\gamma^1$ sets the saddle strength and $\gamma^2$ its asymmetry.

\paragraph{Monkey saddle component.}
A higher-order saddle point is defined as
\begin{equation*}
f_M(x)
=
\tfrac{1}{3}\,\delta^1
\bigl(u^3 - 3uv^2\bigr),
\end{equation*}
where $(u,v)=R_S\tilde{x}$ and $\delta^1$ sets its strength.

\paragraph{Confinement and tilt.}
To keep the function bounded, we add a quartic term
\begin{equation*}
f_{\mathrm{conf}}(x)
=
\tfrac{1}{4}\,c\|\tilde{x}\|^4 .
\end{equation*}
A linear tilt
\begin{equation*}
f_{\mathrm{tilt}}(x)
=
t^\top \tilde{x}
\end{equation*}
is also added to break symmetry.

\paragraph{Random Fourier Feature (RFF) components.}
Each RFF component \cite{Rahimi2007RandomFF} approximates a Gaussian-process sample with an RBF kernel:
\begin{equation*}
f_{\mathrm{RFF}}(x)
=
\sqrt{2}
\sum_{m=1}^{M}
\omega^1_m
\cos\!\bigl((\omega^2_m)^\top x + \omega^3_m\bigr),
\end{equation*}
with phases $\omega^3_m \sim \mathcal{U}(0,2\pi)$, weights
$\omega^1_m \sim \mathcal{N}(0,\sigma^2/M)$, and frequencies
$\omega^2_m \sim \mathcal{N}(\mathbf{0},\ell^{-2}\mathbf{I}_D)$, where $M$ is
the number of \textit{Random Fourier Features}. The lengthscale
$\ell \sim \mathcal{U}(\ell_{\min},\ell_{\max})$ controls smoothness, while
$\sigma \sim \mathcal{U}(\sigma_{\min},\sigma_{\max})$ controls amplitude; the
$1/M$ scaling keeps the variance bounded as $M$ increases. We sample two independent
RFF components of this form. To introduce multi-scale variation, each frequency is
multiplied by a bounded bimodal log-normal jitter factor.

All deterministic geometric components, namely the quadratic, saddle, monkey-saddle,
confinement, and tilt terms, are evaluated on shifted coordinates normalized by the
domain half-width, while the RFF components are evaluated on the original input
coordinates.

To sample a new function, we sample the mixing weights $(\alpha_1, \dots, \alpha_5)$, function type, geometric parameters $(\beta, \gamma, \delta, R, R_S)$, and RFF parameters $(\ell_j, \sigma_j, \omega^{j})$ for each GP components. The resulting objectives vary naturally in topology, from smooth bowls to saddle points and multi-modal landscapes, reflecting the diversity induced by the prior rather than any manual shaping of function landscapes. Samples of our prior are visually presented in \Cref{fig:prior}.

\subsection{Parameter values of our proposed Prior}
\label{appx:prior_params}
\begin{table}[h]
\centering
\caption{Main hyperparameter values of the proposed prior.}
\begin{tabular}{ll}
\toprule
Symbol / Quantity & Distribution / Value \\
\midrule
$D$ & $2$ \\
Domain & $[-50,50]^D$ \\
Input shift & $\mathcal{U}(-40,40)^D$ \\
Vertical offset $b$ & $0$ \\
$M$ & $1000$ \\
Quadratic mixing weight $\alpha_3$ & $\mathcal{U}(0.05,0.4)$ for bowl-dominated instances \\
Saddle strength $\gamma^1$ & log-$\mathcal{U}(1,12)$ \\
Saddle asymmetry $\gamma^2$ & $\mathcal{U}(0.3,2.5)$ \\
Monkey-saddle strength $\delta^1$ & log-$\mathcal{U}(0.5,6)$ \\
Confinement coefficient $c$ & log-$\mathcal{U}(0.5,3)$ \\
Tilt strength $\|t\|$ & $\mathcal{U}(0,0.5)$ \\
RFF kernel & RBF \\
RFF lengthscale $\ell$ & $\mathcal{U}(4,10)$ \\
RFF amplitude $\sigma$ & $\mathcal{U}(1,5)$ \\
Quadratic curvature scale $\beta^1_d$ & $\mathcal{U}(0.5,8.0)$ \\
Rotation probability & $0.7$ \\
Ill-conditioning probability & $0.5$ for bowl-dominated instances, $0.4$ otherwise \\
Condition number $\kappa$ & $\mathcal{U}(10^3,10^6)$ for bowl-dominated instances, $\mathcal{U}(10^1,10^4)$ otherwise \\
\bottomrule
\end{tabular}
\label{05:prior_values}
\end{table}

The five mixing weights $(\alpha_1,\ldots,\alpha_5)$ are sampled as normalized positive uniform weights and then adjusted according to the sampled landscape type. In particular, saddle-dominated instances receive a non-negligible saddle weight, monkey-saddle terms are removed when inactive, and the remaining active weights are renormalized. The table reports only the main hyperparameters; implementation-level details, including the exact landscape-type adjustment rules and frequency-jitter parameters, are provided in the published code.

\subsection{Visual Illustration of the Prior}
\label{appx:prior_viz}

\clearpage

%% file: 09_appendix_c.tex
\section{Architecture and Training Details}
\label{app_c}

This appendix reports the architectural and training hyperparameters used in all experiments. Unless stated otherwise, a single configuration is used across ablations and evaluations to isolate the effect of the proposed optimizer design from implementation-specific tuning.
\Cref{tab:hyperparams} summarizes the transformer-based architecture shared between actor and critic, while \Cref{tab:ppo_hparams} details the PPO training setup and optimization hyperparameters. We set the boundary target scale as \( V_x = V_y = 3 \). Together, these settings fully specify the model and training procedure required to reproduce the reported results.

\begin{table}[h]
\centering
\caption{Key architecture hyperparameters for the shared Transformer backbone and actor/critic heads.}
\small
\begin{tabular}{ll}
\hline
\textbf{Component} & \textbf{Value} \\
\hline
\multicolumn{2}{l}{\textit{Shared backbone (Transformer)}} \\
Embedding dimension & 64 \\
\# Transformer blocks & 8 \\
\# Attention heads & 8 ($d_{\text{head}} = 8$) \\
Dropout & 0.05 \\
Pooling & Last-token pooling \\
Shared output dim & 32 \\
\multicolumn{2}{l}{\textit{Actor / Critic heads}} \\
Actor head (MLP) & $32 \rightarrow 256 \rightarrow 256 \rightarrow 1$ \\
Critic head (MLP) & $32 \rightarrow 256 \rightarrow 256 \rightarrow 1$ \\
\hline
\end{tabular}
\label{tab:hyperparams}
\end{table}

\begin{table}[h]
\centering
\caption{PPO training hyperparameters and optimizer settings.}
\small
\begin{tabular}{ll}
\hline
\textbf{Component} & \textbf{Value} \\
\hline
Actor lr & $5\times 10^{-5}$ \\
Critic lr & $5\times 10^{-5}$ \\
Optimizer & \texttt{AdamWScheduleFree \cite{defazio2024road}} \\
Actor weight decay & $1\times 10^{-4}$ \\
Critic weight decay  & $1\times 10^{-4}$ \\
Actor warmup & 300 \\
Critic warmup & 10 \\
PPO clip ratio & 0.1 \\
GAE $\gamma$  &  0.97 \\
GAE $\lambda$ &  0.9 \\
Epochs over collected trajectories &  4 \\
PPO Minibatch size &  8192 \\
Gradient clip norm & 0.5 \\
KL early stopping & 0.01 \\
Adaptive KL penalty & yes (initial coef 0.1, target KL 0.01) \\
Return normalization &  Online Welford estimator \\
\hline
\end{tabular}
\label{tab:ppo_hparams}
\end{table}

\newpage

%% file: 10_appendix_d.tex
\section{Baseline Hyperparameter Optimization and Qualitative Analysis}
\label{app_d}

This appendix contains additional experimental details that support the evaluation in \Cref{sec:evaluation}. It is divided into two parts. First, we report the hyperparameter optimization (HPO) procedure used for baseline methods, including the learning-rate sweeps and selected configurations. These results ensure that all baselines are evaluated under competitive and well-tuned settings.  

\begin{figure}
\begin{subfigure}{0.5\textwidth} 
    \centering
    \includegraphics[width=\textwidth]{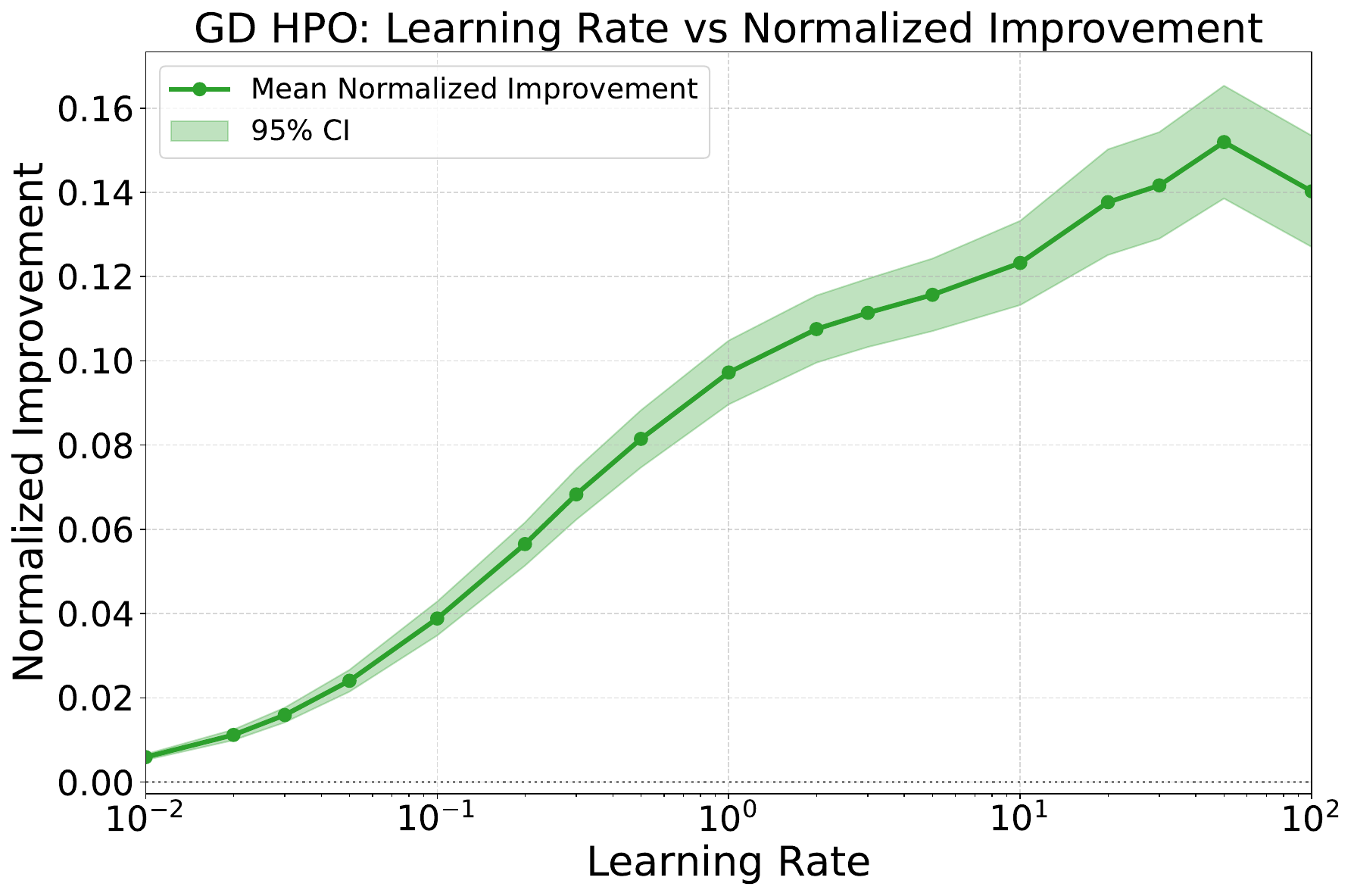}
    \caption{GD peaks at a learning rate of 50.}
    \label{fig:hyp_sgd_regret}
\end{subfigure}
\begin{subfigure}{0.5\textwidth} 
    \centering
    \includegraphics[width=\textwidth]{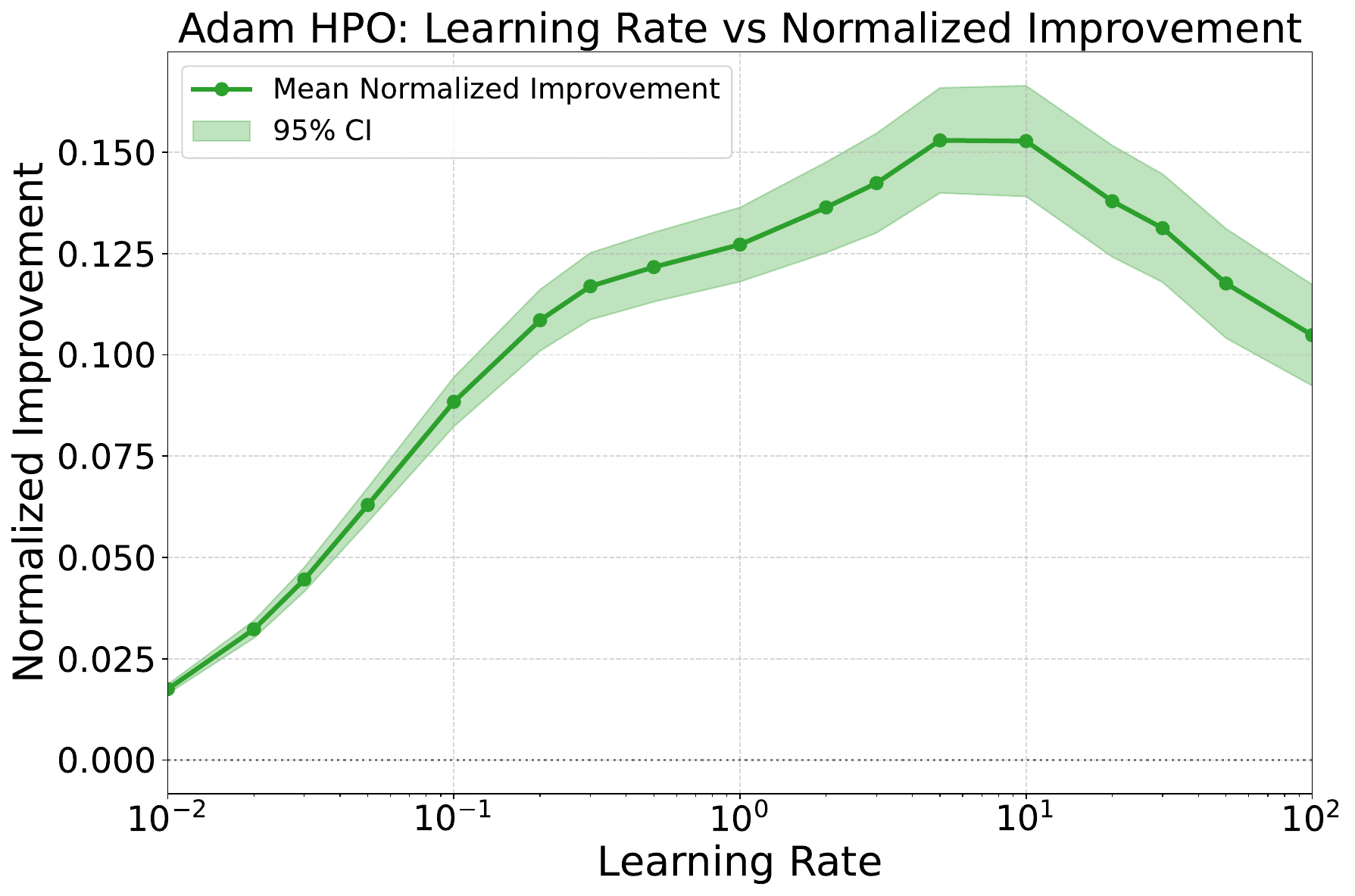}
    \caption{Adam peaks at a learning rate of 10.}
    \label{fig:hyp_adam_regret}
\end{subfigure}\\
\begin{subfigure}{0.5\textwidth}
    \centering
    \includegraphics[width=\textwidth]{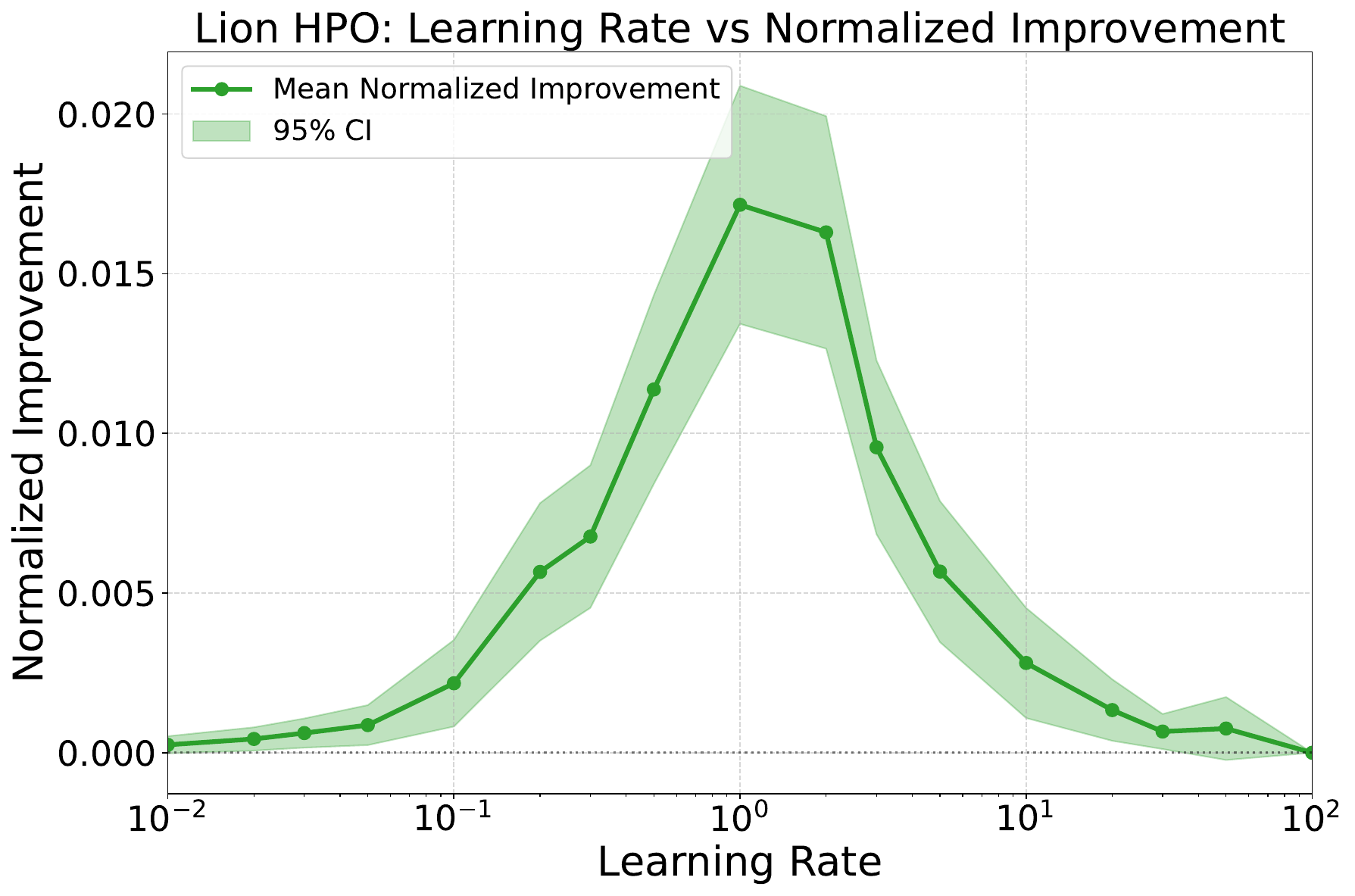}
    \caption{Lion peaks at a learning rate of 1.}
    \label{fig:hyp_lion_regret}
\end{subfigure}
\begin{subfigure}{0.5\textwidth} 
    \centering
    \includegraphics[width=\textwidth]{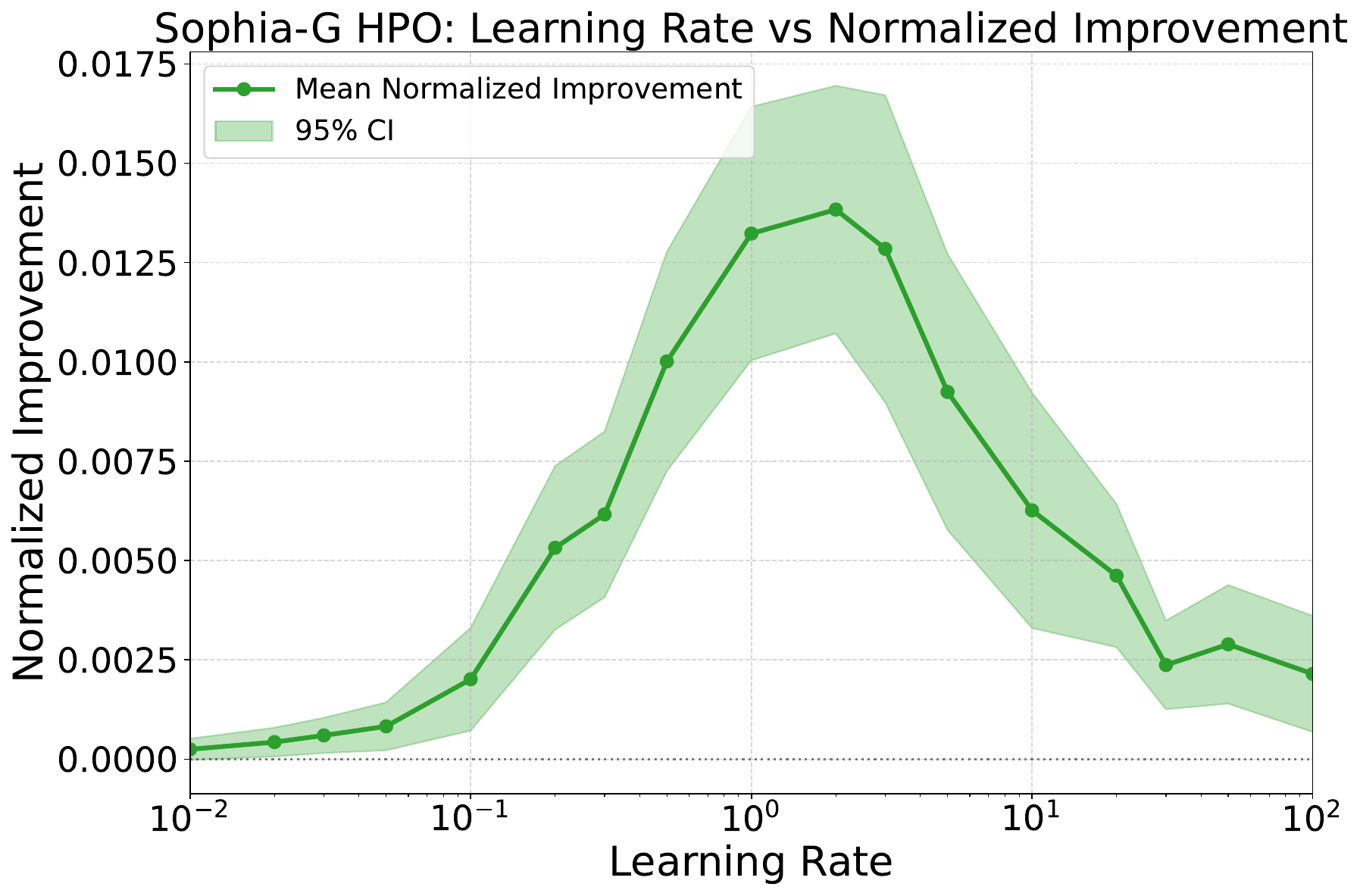}
    \caption{Sophia peaks at a learning rate of 2.}
    \label{fig:hyp_sophia_regret}
\end{subfigure}\\
\begin{subfigure}{1\textwidth} 
    \centering
    \includegraphics[width=0.5\textwidth]{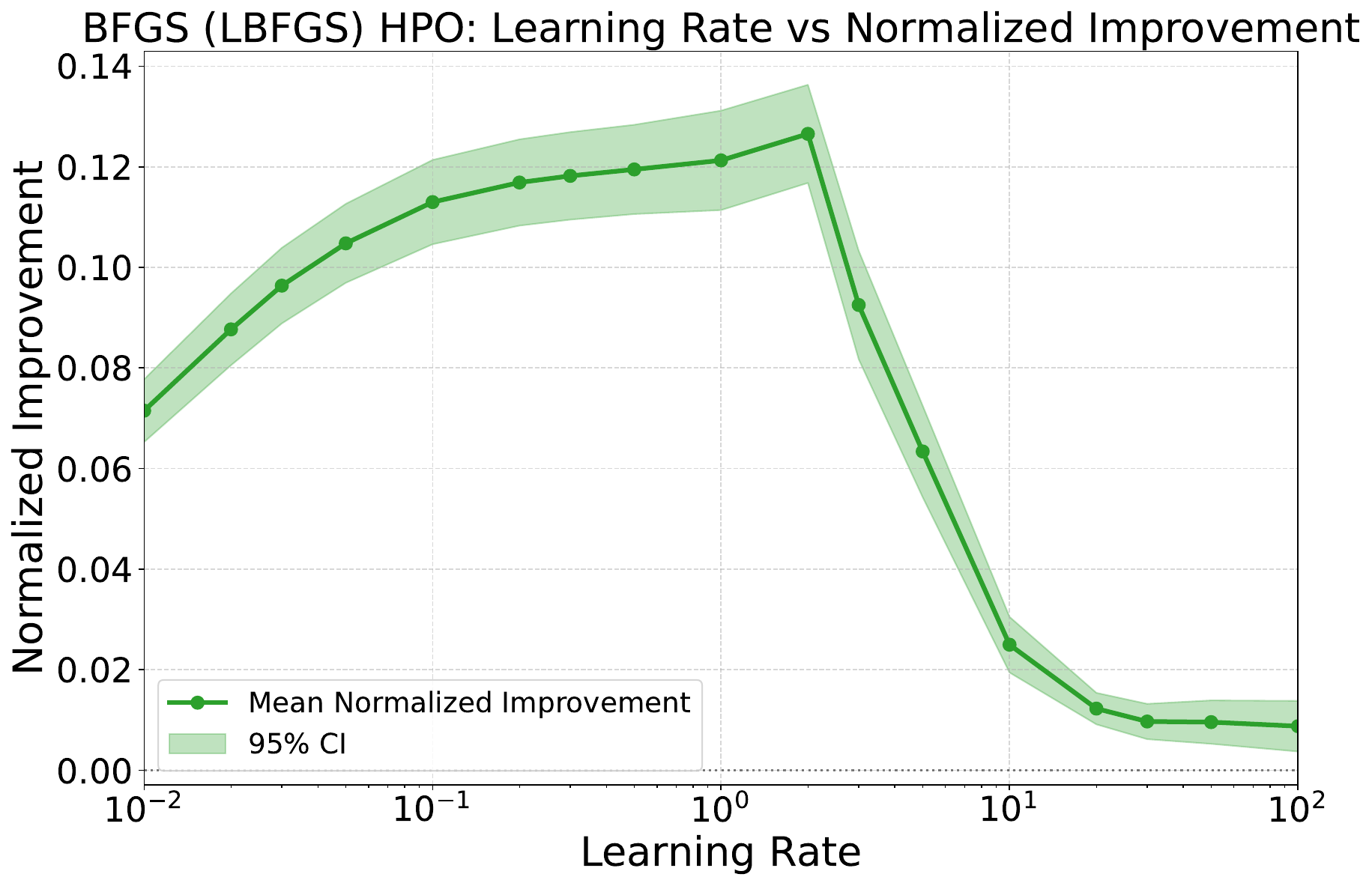}
    \caption{L-BFGS peaks at a learning rate of 2.}
    \label{fig:hyp_bfgs_regret}
\end{subfigure}
\caption{Learning rate sweep on the in-distribution validation set. }
\end{figure}

We sweep the initial learning rate for GD, Adam, Lion, Sophia-G and L-BFGS using a grid search over
\[
\eta \in \{0.01, 0.02, 0.03, 0.05, 0.1, 0.2, 0.3, 0.5, 1.0, 2.0, 3.0, 5.0, 10.0, 20.0, 30.0, 50.0, 100.0\}.
\]
All other hyperparameters are kept at their default values. For L-BFGS, this implies that the optimizer is allowed to perform more function evaluations per optimization iteration than the other baselines.  
\Cref{fig:hyp_sgd_regret}, \Cref{fig:hyp_adam_regret},\Cref{fig:hyp_lion_regret}, \Cref{fig:hyp_sophia_regret} and \Cref{fig:hyp_bfgs_regret} show the learning-rate sweeps for GD, Adam, Lion, Sophia, and L-BFGS on the validation set, respectively.

For GA and DE, we sweep the crossover rate and mutation factor using a grid search over
\[
crossover\_rate \in \{0.1, 0.3, 0.5, 0.7, 0.9\} \quad
mutation\_factor \in \{0.1, 0.3, 0.5, 0.7, 0.9\}.
\]

For CMAES, we sweep the initial step size using a grid search over

\[
init\_step\_size \in \{0.2, 0.3, 0.4, 0.5, 0.6, 0.7, 0.8, 0.9, 1.0, 2.0, 3.0, 4.0, 5.0, 10.0, 20.0\}.
\]

\begin{figure}
\begin{subfigure}{0.5\textwidth} 
    \centering
    \includegraphics[width=\textwidth]{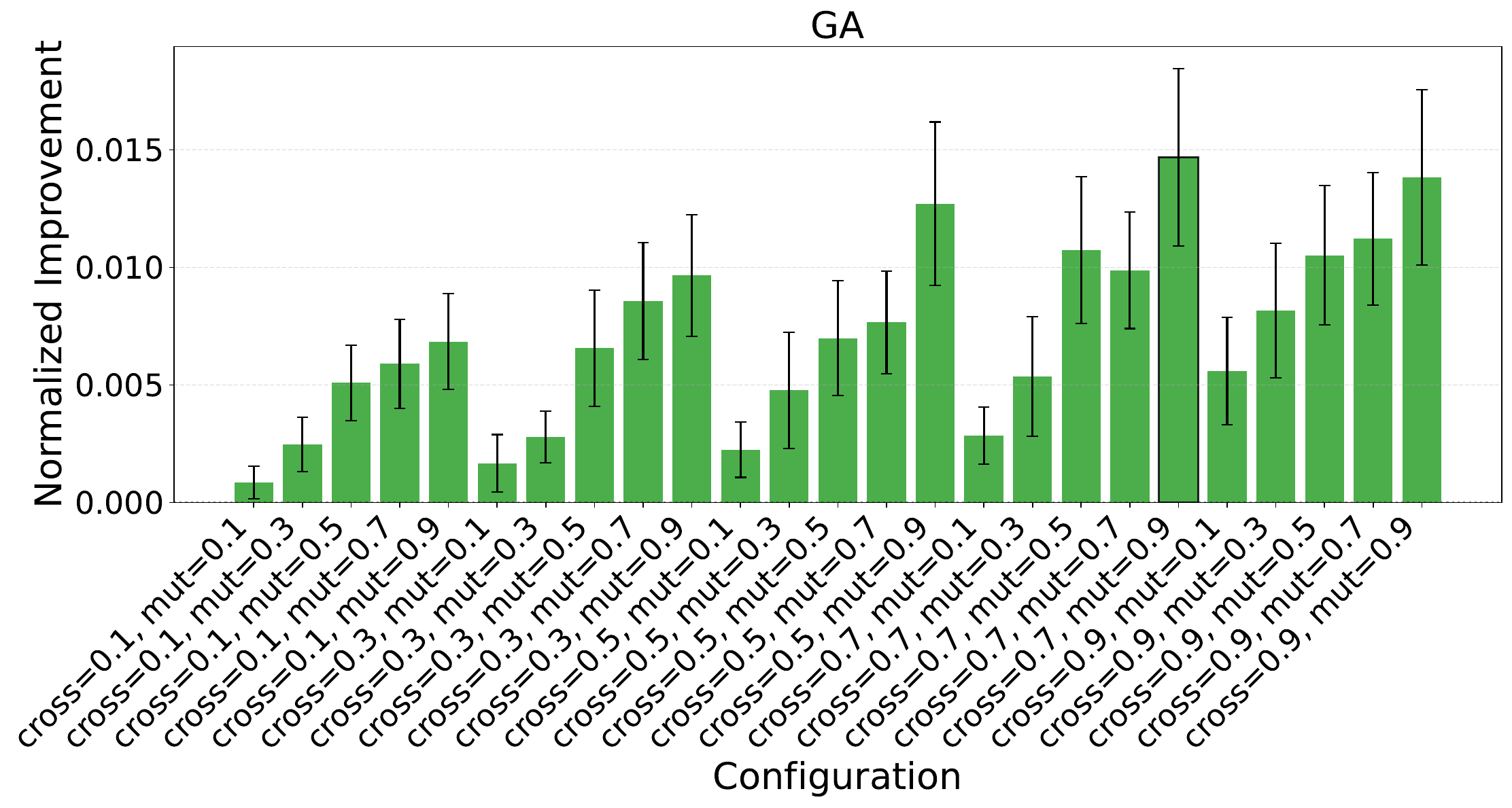}
    \caption{GA peaks at a crossover rate of 0.7 and mutation factor of 0.9.}
    \label{fig:hyp_ga_regret}
\end{subfigure}
\begin{subfigure}{0.5\textwidth}
    \centering
    \includegraphics[width=\textwidth]{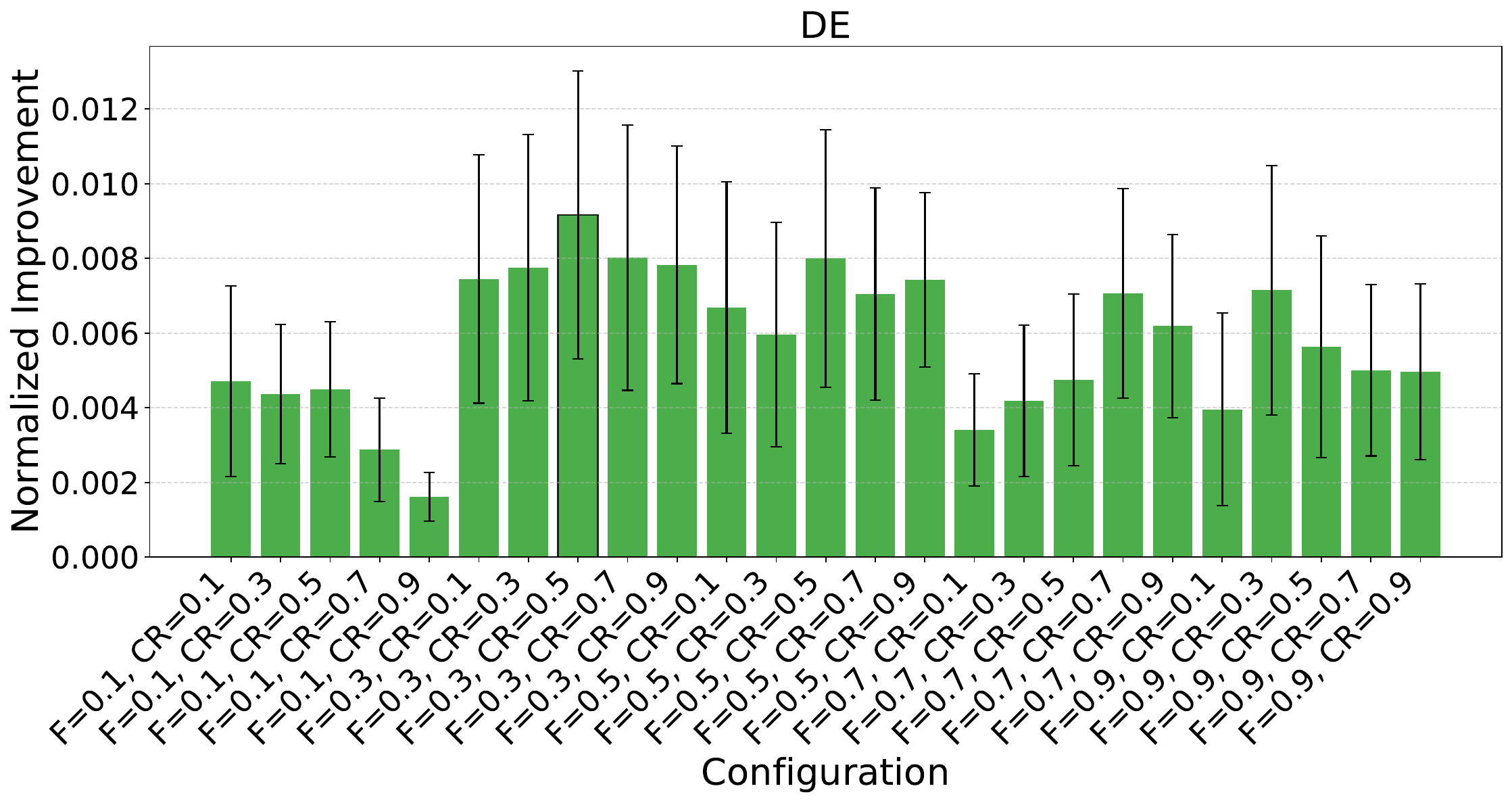}
    \caption{DE peaks at a crossover rate of 0.5 and mutation factor of 0.3.}
    \label{fig:hyp_de_regret}
\end{subfigure}\\
\begin{subfigure}{0.5\textwidth} 
    \centering
    \includegraphics[width=\textwidth]{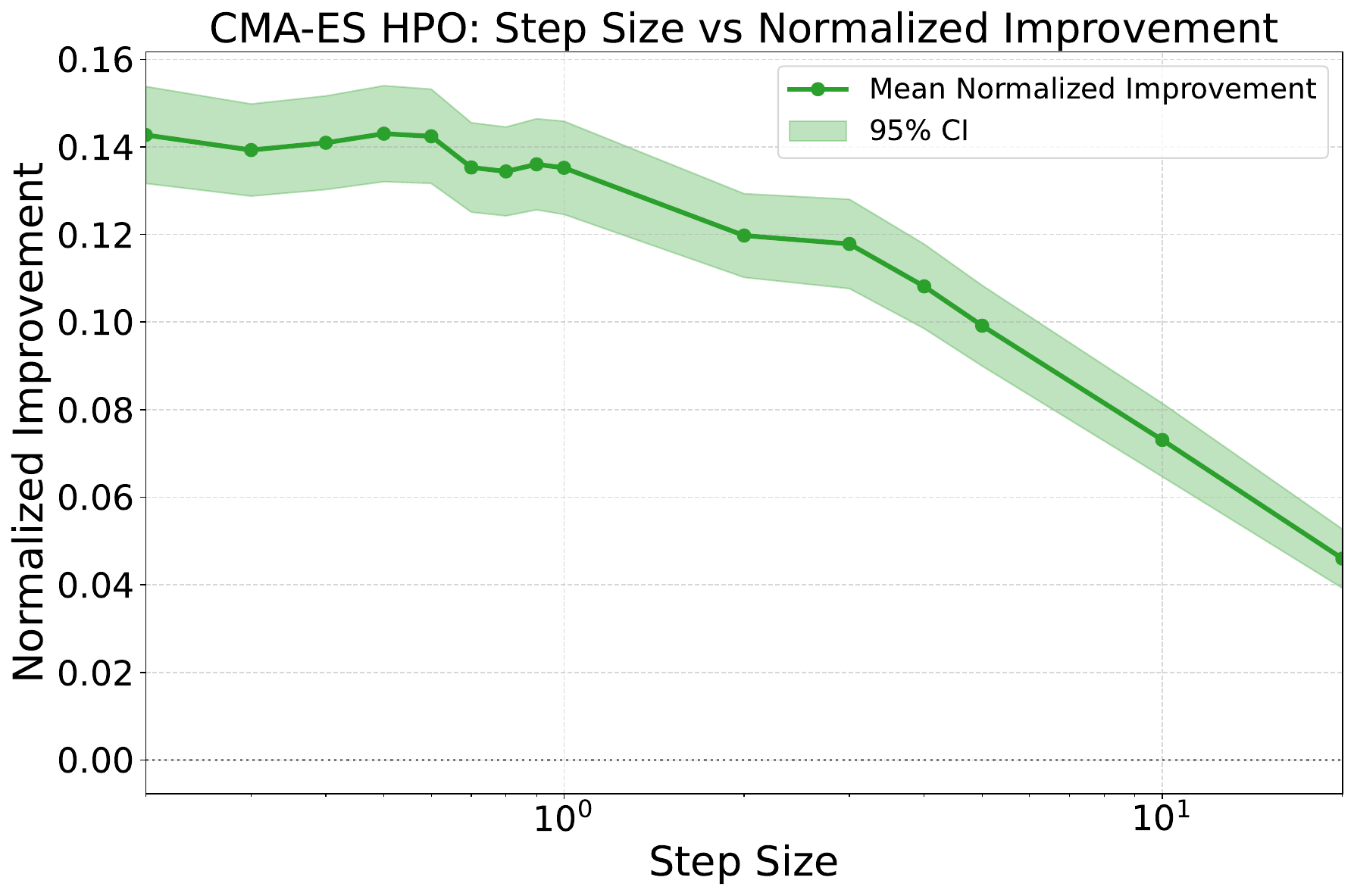}
    \caption{CMAES peaks at an initial step size of 0.2.}
    \label{fig:hyp_cmaes_regret}
\end{subfigure}
\caption{Hyperparameter sweep for the evolutionary methods on the in-distribution validation set. }
\end{figure}

We keep the population size at 10 for all evolutionary methods.
\Cref{fig:hyp_ga_regret}, \Cref{fig:hyp_de_regret} and \Cref{fig:hyp_cmaes_regret} show the sweeps for GA , DE and CMAES on the validation set, respectively.

\clearpage

\subsection{In-Distribution Performance}
In the second part of this appendix, we provide qualitative visualizations of optimization trajectories produced by POP on functions sampled from the synthetic prior. These figures complement the quantitative results reported in \Cref{sec:evaluation} by illustrating typical optimization behaviors observed in practice.

They highlight how POP adapts its step sizes and exploration behavior in response to the local geometry of the objective function.

\begin{figure}[h]
    \centering
    \includegraphics[width=0.8\textwidth, height=5cm]{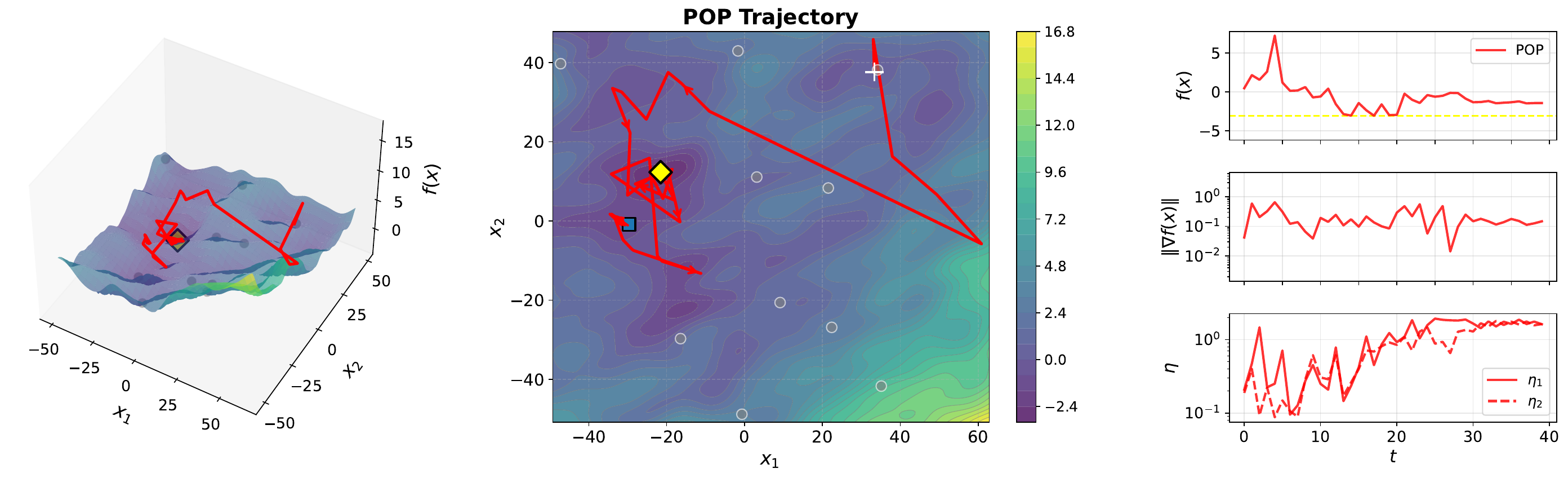}

        \includegraphics[width=\textwidth, height=5cm]{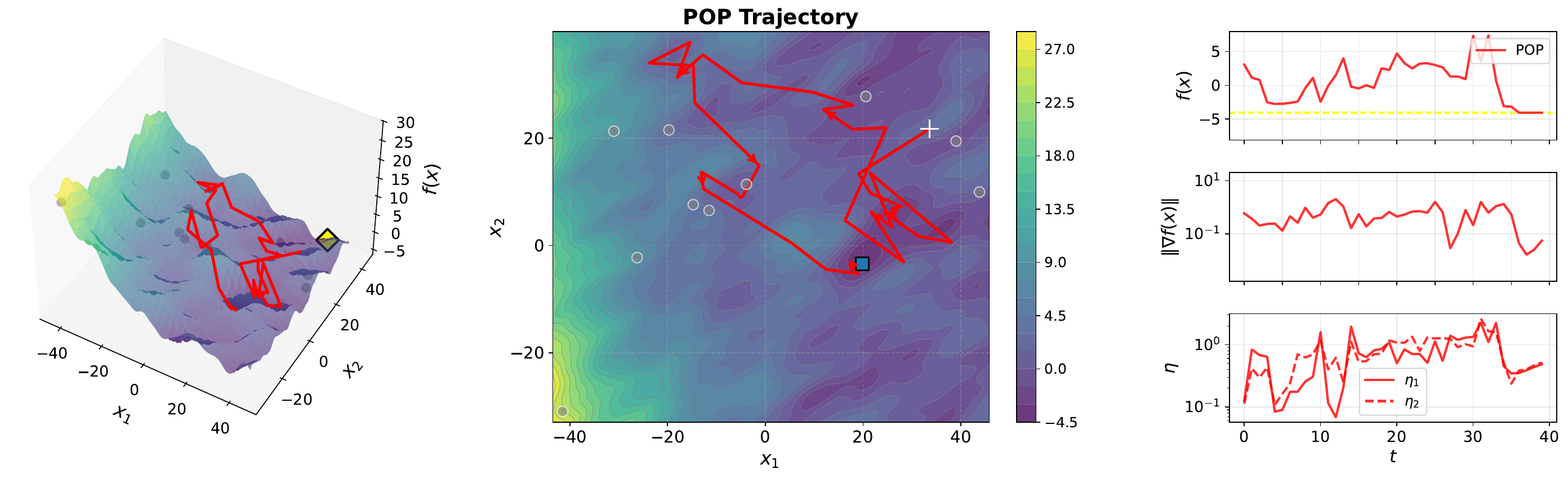}
            \includegraphics[width=\textwidth, height=5cm]{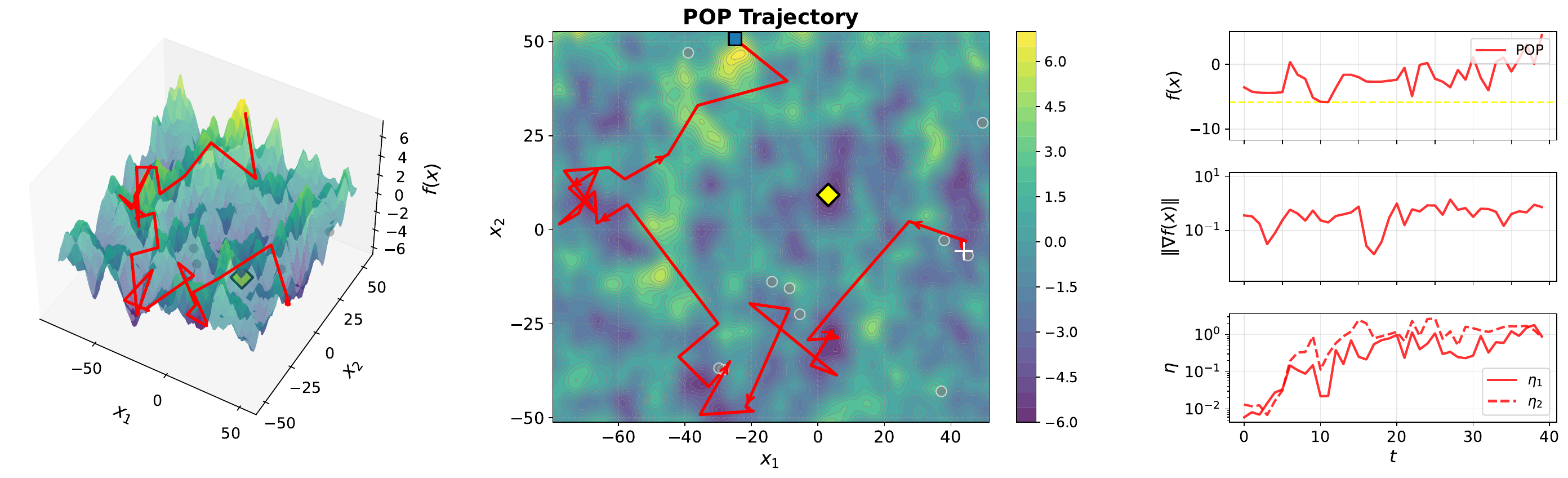}

    \caption{Trajectory of POP's on in-distribution functions. The algorithm progresses from the initial state (white cross) terminating at the final state (square).}
\label{example_convex}
\end{figure}

\label{appd:in_dist}
\begin{figure}[h]
    \centering

    \includegraphics[width=0.8\textwidth, height=5cm]{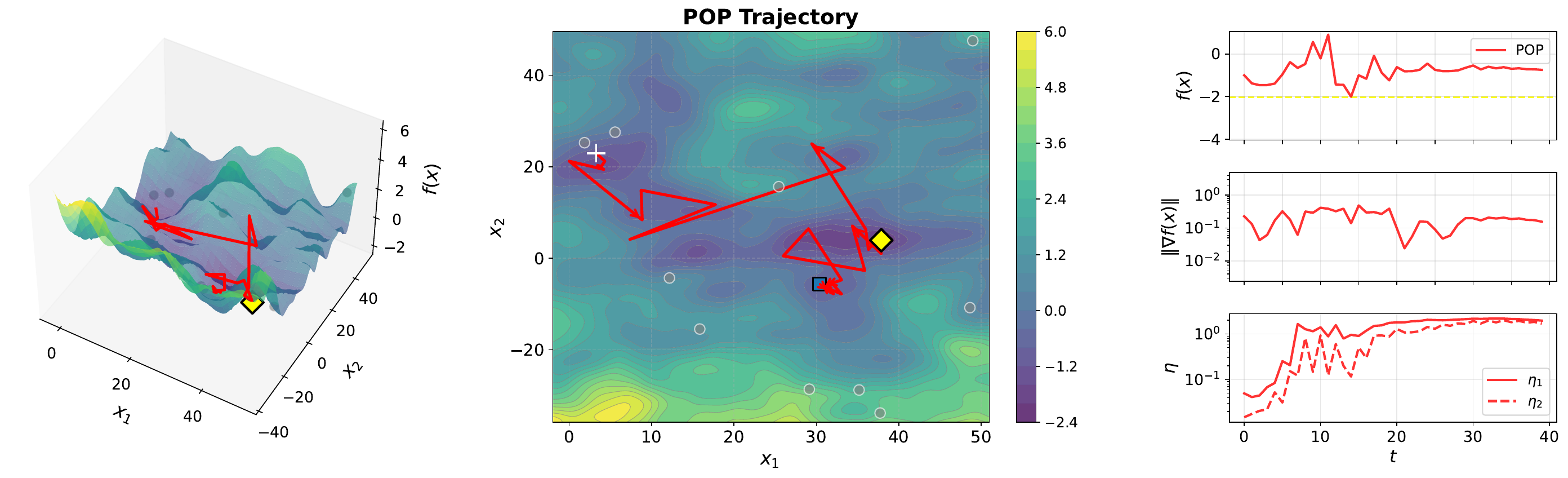}
        \includegraphics[width=\textwidth, height=5cm]{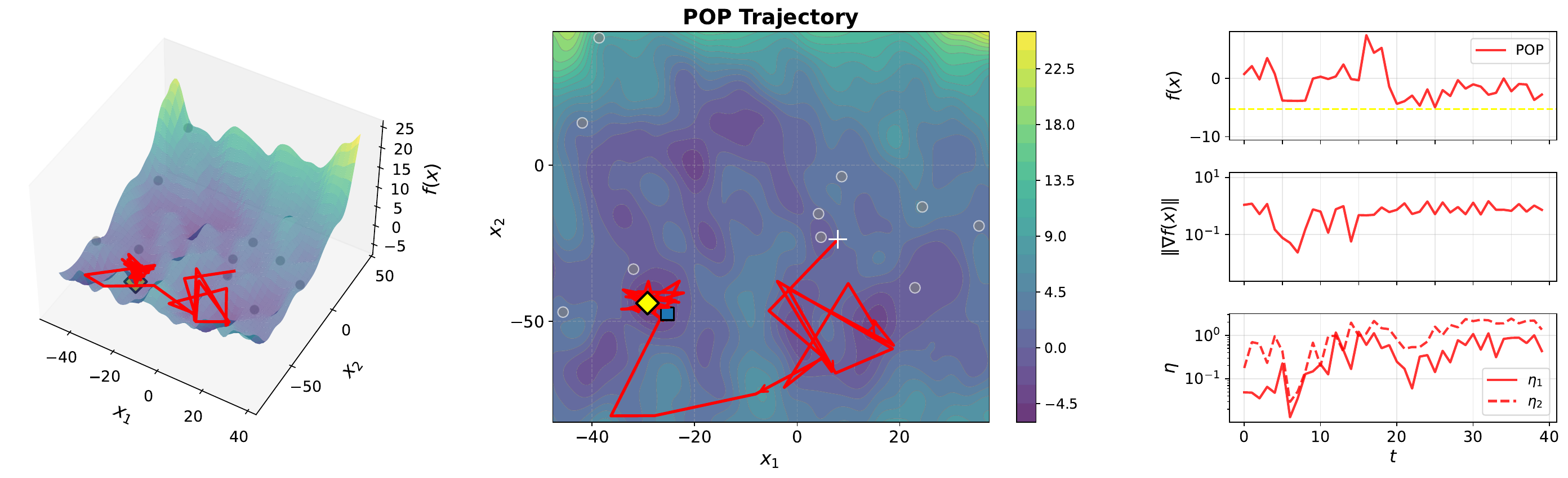}
    \caption{Trajectory of POP's on in-distribution functions. The algorithm progresses from the initial state (white cross) terminating at the final state (square).}
\label{example_non_convex}
\end{figure}


\clearpage

%% file: 10_appendix_h.tex
\section{Extended Comparison Against Optimization Solvers}
\label{app_h}

For the extended comparison we compare our method against a diverse set of optimization baselines. \textbf{Gradient Descent (GD)} \citep{cauchy1847},  \textbf{Adam} \citep{Kingma_2014_adam}, and \textbf{Lion} \citep{chen2023symbolic} represent first-order gradient-based optimizers, while \textbf{Sophia-G} \citep{liu2023sophia} is a second-order-inspired optimizer that uses a lightweight diagonal Hessian approximation. \textbf{L-BFGS} \citep{liu1989L-bfgs} is a quasi-Newton method leveraging approximate second-order information. \textbf{Random Search} provides a simple derivative-free baseline, while \textbf{Genetic Algorithms} \citep{holland1975adaptation}, \textbf{Differential Evolution} \citep{storn1997differential}, and \textbf{CMA-ES} \citep{hansen2016cma} are population-based evolutionary methods for global optimization. \textbf{TPE} \citep{bergstra2011algorithms}, \textbf{GP}-based Bayesian optimization \citep{Rasmussen2006Gaussian}, and \textbf{HEBO} \citep{cowen2022hebo} are surrogate-based Bayesian optimization methods. Finally, \textbf{VeLo} \citep{Metz_2022_VeLO} and \textbf{PFNs4BO} \citep{muller2023pfns4bo} are learned optimization baselines: VeLo learns an optimizer from a prior over neural network training tasks, while PFNs4BO uses a prior-data fitted network as a learned surrogate for Bayesian optimization. All baselines are evaluated under the same optimization budget.

\subsection{Hypothesis 1: Extended comparison} \label{app_h:h1}
\begin{figure}[H]
    \includegraphics[width=0.7\linewidth]{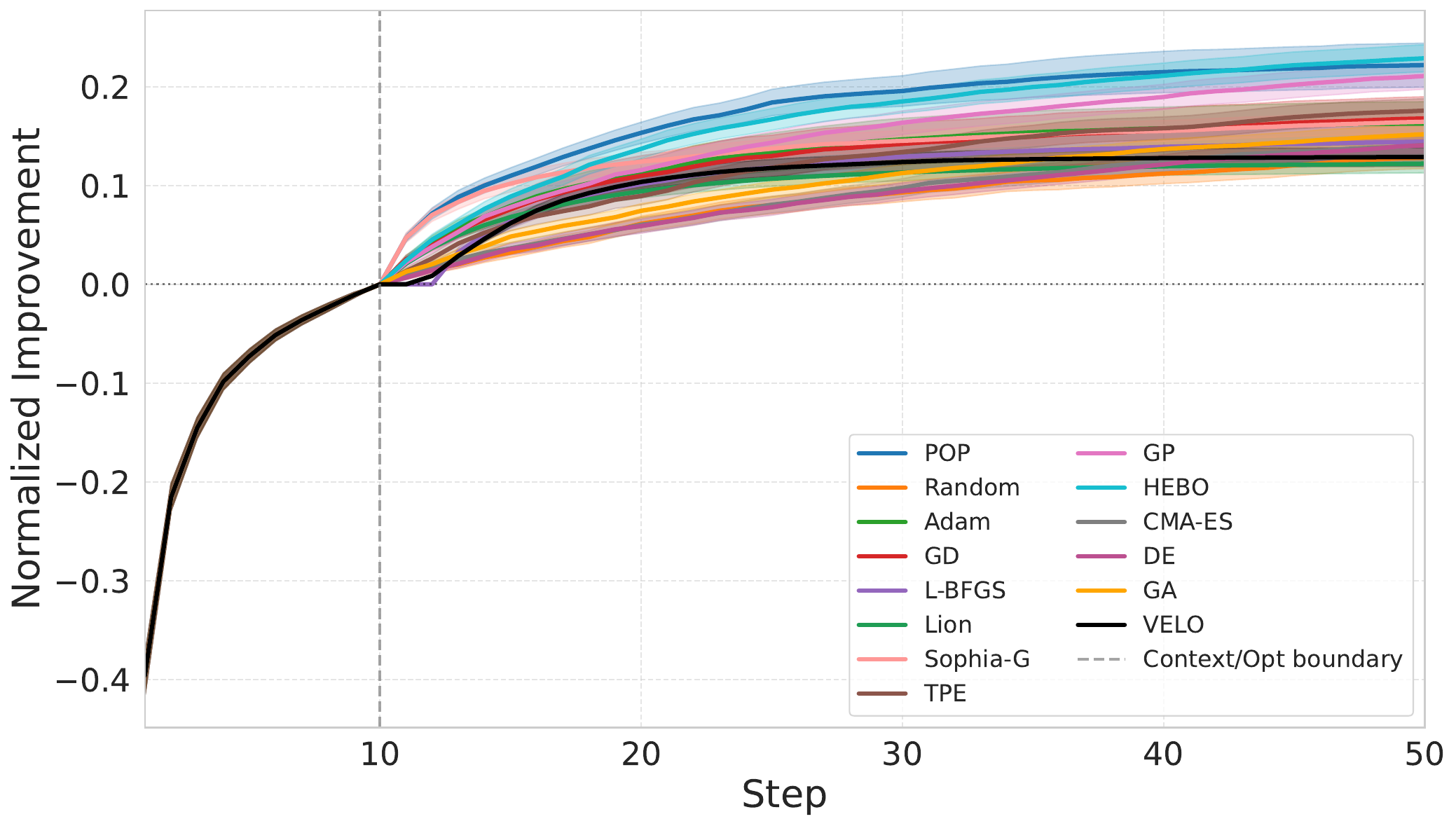}
    \captionof{figure}{In-distribution test set performance vs. all baselines. Mean normalized improvement over steps; shading indicates 95\% CIs. Dashed line marks the context/optimization boundary.}
\end{figure}

\subsection{Hypothesis 2: Extended comparison} \label{app_h:h2}

\begin{figure}[H]
    \includegraphics[width=0.7\linewidth]{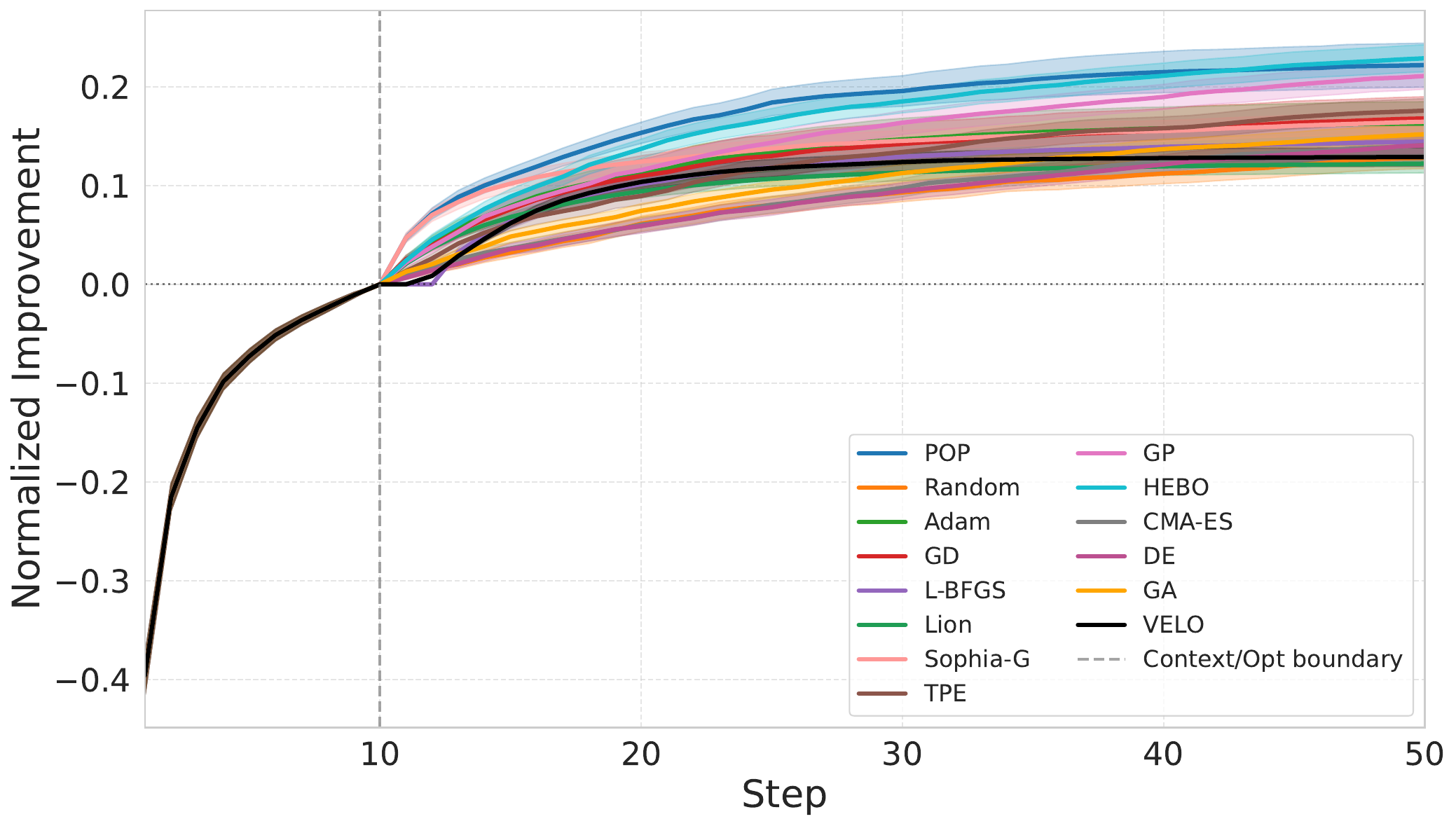}
    \captionof{figure}{In-distribution test set performance vs. all baselines. Mean normalized improvement over steps; shading indicates 95\% CIs. Dashed line marks the context/optimization boundary.}
\end{figure}

\begin{table}[H]
\centering
\caption{Mean performance across different dimensionalities.}
\resizebox{\columnwidth}{!}{
\begin{tabular}{lcccccccccccccc}
\hline
Dim. & POP & Random & Adam & GD & L-BFGS & Lion & Sophia-G & TPE & GP & HEBO & CMA-ES & DE & GA & VELO \\
\hline
16D & 0.2911 & 0.1177  &  0.2887 & 0.2510 & 0.1728 & 0.2409 & 0.2579 & 0.2107 & 0.2692 & 0.2702 & 0.1153 & 0.1421 & 0.2145 & \textbf{0.3100} \\
32D & 0.4966  & 0.1669 & 0.5029 & 0.4003 & 0.4171 & 0.4245 & 0.4820 & 0.3176 & 0.4807 & 0.4368 & 0.0572 & 0.1953 &0.3691 & \textbf{0.5225}  \\
\hline
\end{tabular}
}
\label{tab:dimensionality_results}
\end{table}%

\subsection{Hypothesis 3: Extended comparison}\label{app_h:h3}

\begin{figure}[H]
    \includegraphics[width=0.8\linewidth]{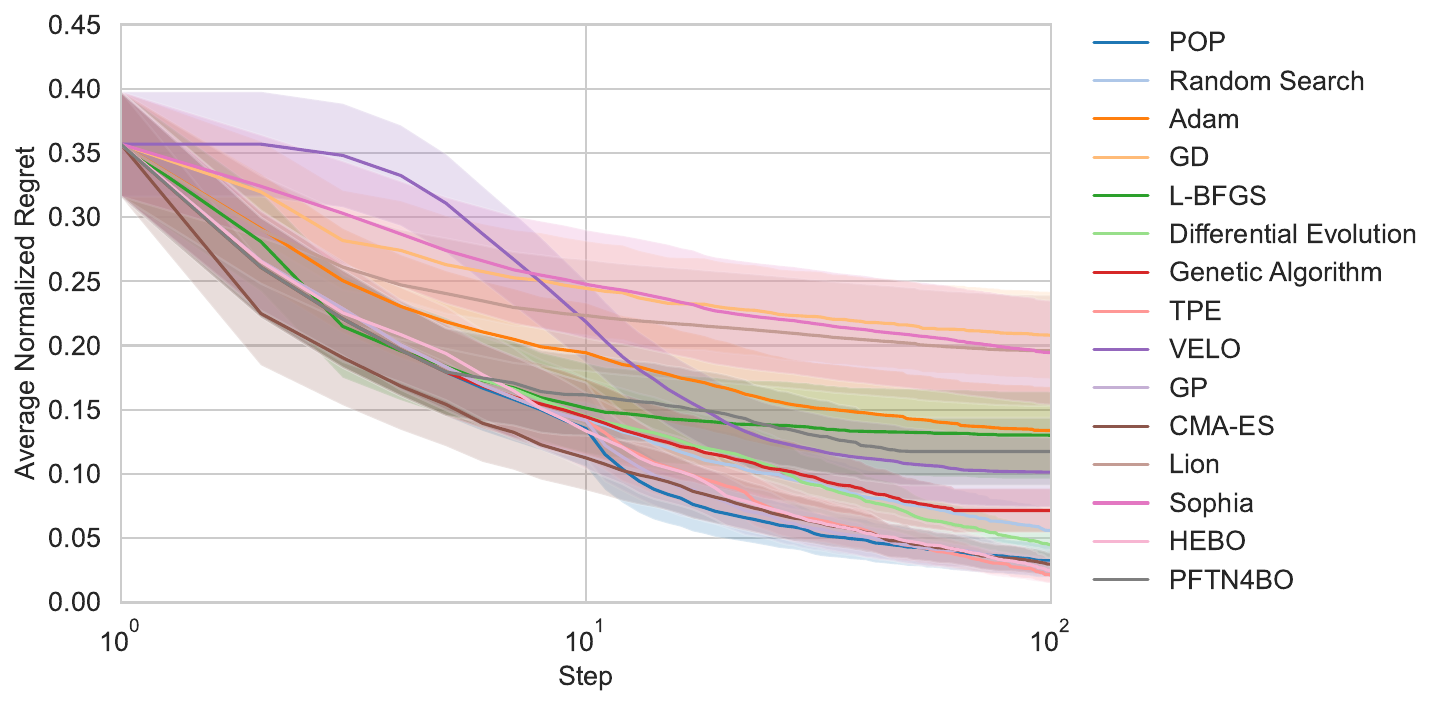}
    \captionof{figure}{The average normalized regret for all methods on VLSE. Solid lines represent the mean value.}
\end{figure}

\begin{figure}
    \centering
    \begin{subfigure}{0.495\textwidth}
        \centering
        \includegraphics[width=\textwidth, trim={0 0 0 1cm}, clip]{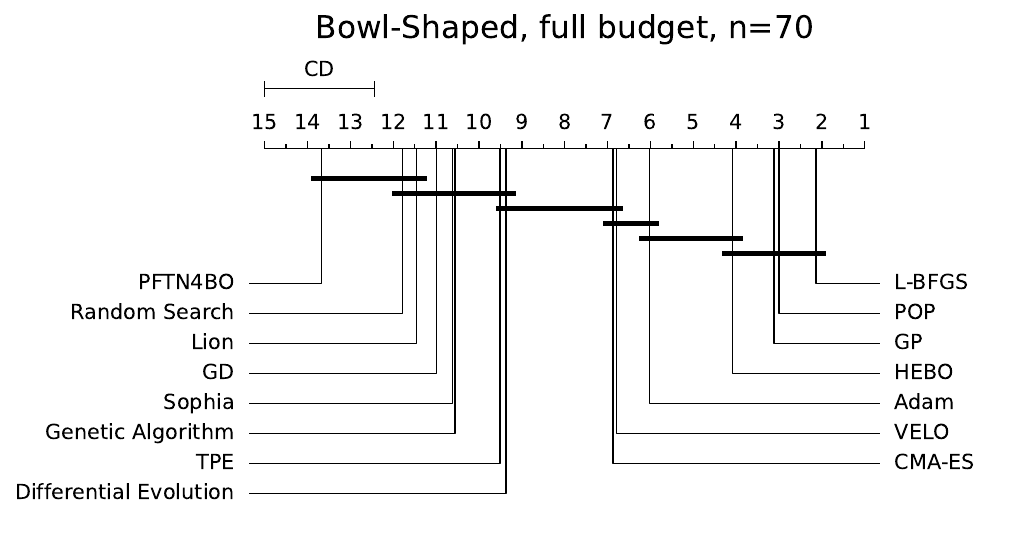}
        \caption{Bowl-shaped}
        \label{subfig:bowlShaped}
    \end{subfigure}
    \begin{subfigure}{0.495\textwidth}
        \centering
        \includegraphics[width=\textwidth, trim={0 0 0 1cm}, clip]{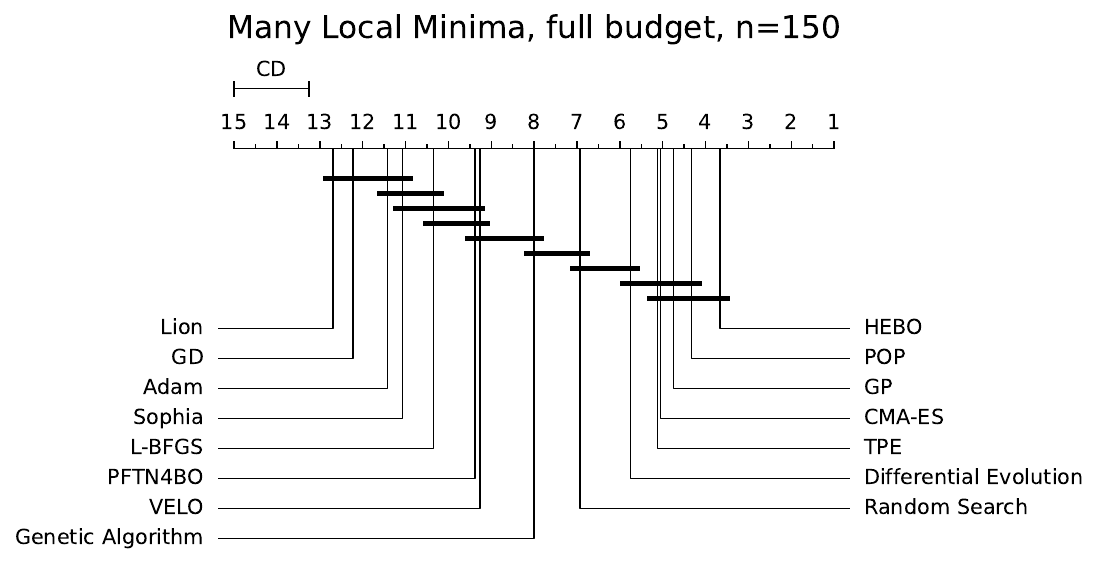}
        \caption{Many Local minima}
        \label{subfig:localMinima}
    \end{subfigure}\\
    \begin{subfigure}{0.495\textwidth}
        \centering
        \includegraphics[width=\textwidth, trim={0 0 0 1cm}, clip]{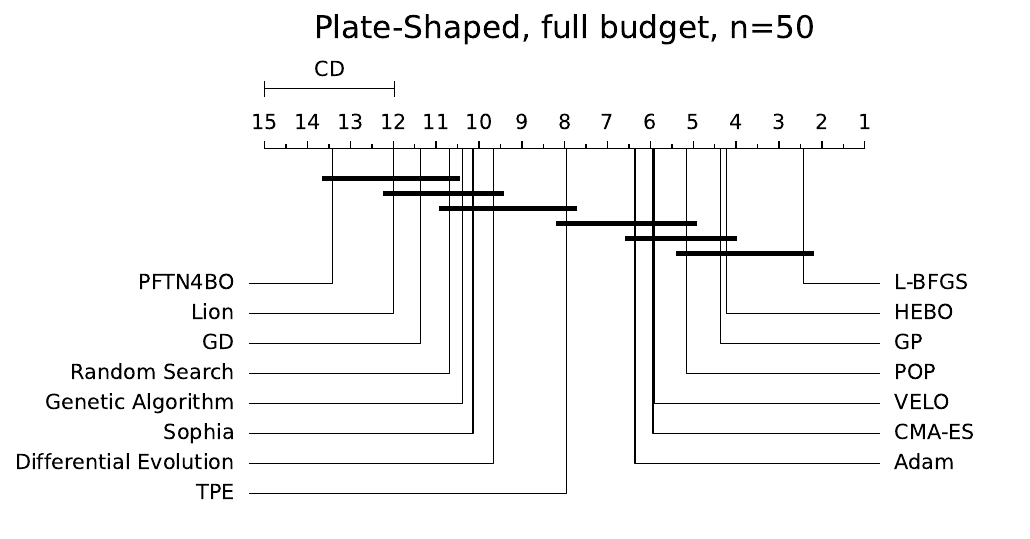}
        \caption{Plate-shaped}
        \label{subfig:plateShaped}
    \end{subfigure}
    \begin{subfigure}{0.495\textwidth}
        \centering
        \includegraphics[width=\textwidth, trim={0 0 0 1cm}, clip]{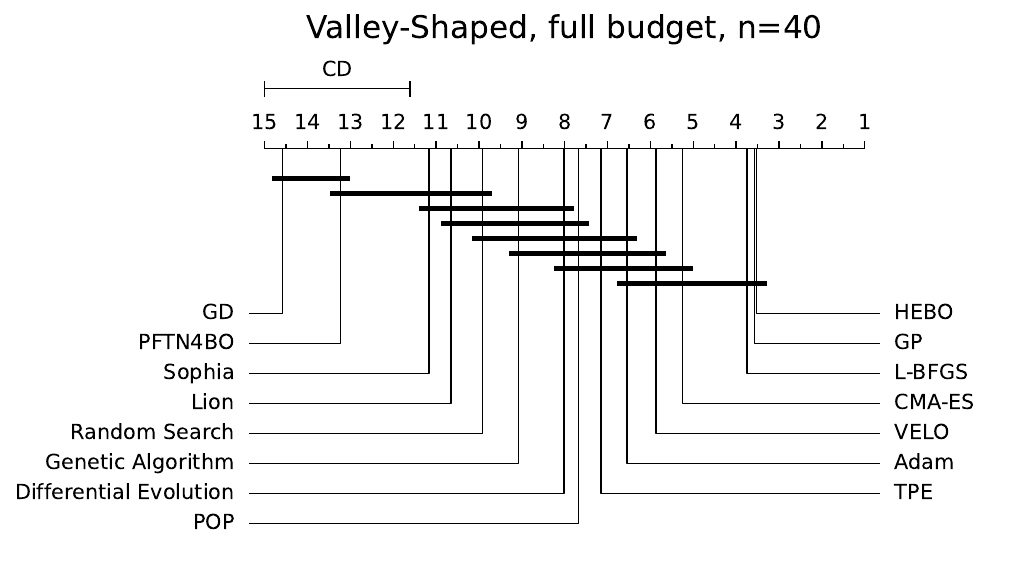}
        \caption{Valley shaped}
        \label{subfig:valleyShaped}
    \end{subfigure}\\
    \begin{subfigure}{\textwidth}
        \centering
        \includegraphics[width=0.6\textwidth, trim={0 0 0 1cm}, clip]{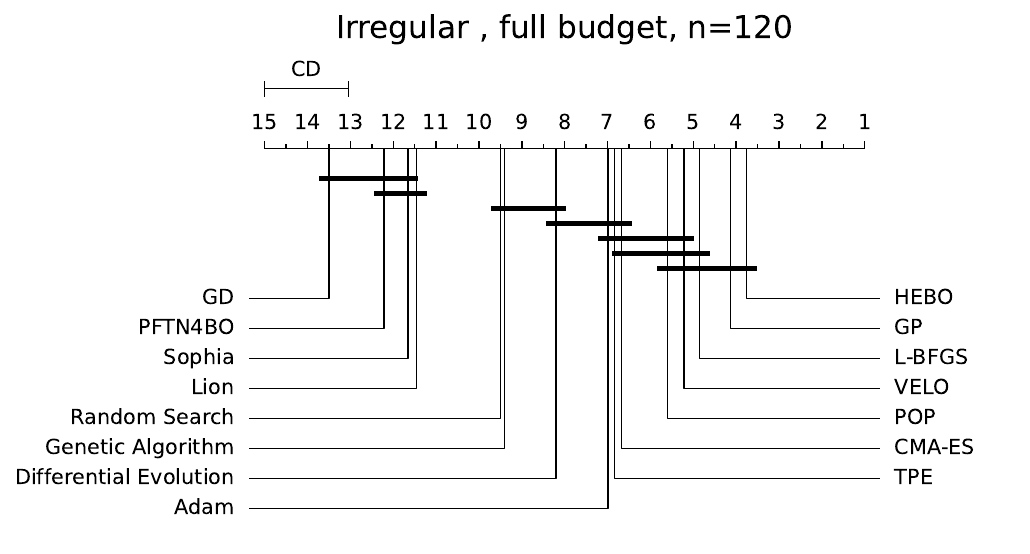}
        \caption{Irregular}
        \label{subfig:irregular}
    \end{subfigure}\\
    \caption{Performance of POP compared to individual competitors from different model categories on diverse function classes}
    \label{fig:ood_categories}
\end{figure}

\Cref{fig:ood_categories} shows the performance of POP against individual competitors of different model families on diverse function classes according to the VLSE benchmar dataset~\citep{simulationlib}. 

\begin{figure}
    \centering
    \begin{subfigure}{0.495\textwidth}
        \centering
        \includegraphics[width=\textwidth, trim={0 0 0 1cm}, clip]{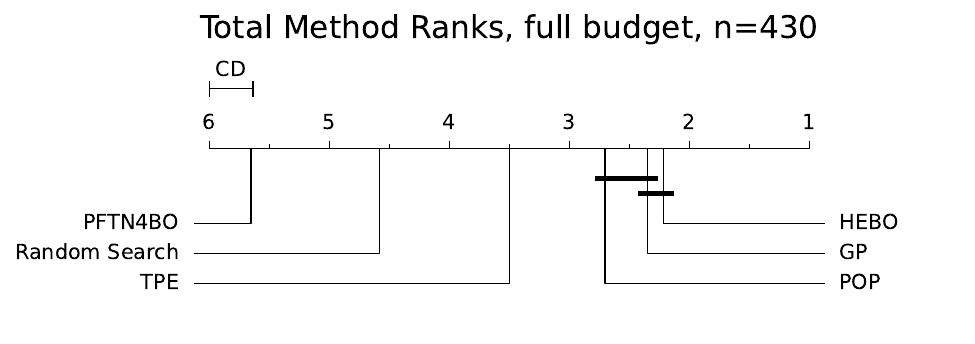}
        \caption{POP vs. Bayesian Optimization}
        \label{subfig:pop_vs_BO}
    \end{subfigure}
    \begin{subfigure}{0.495\textwidth}
        \centering
        \includegraphics[width=\textwidth, trim={0 0 0 1cm}, clip]{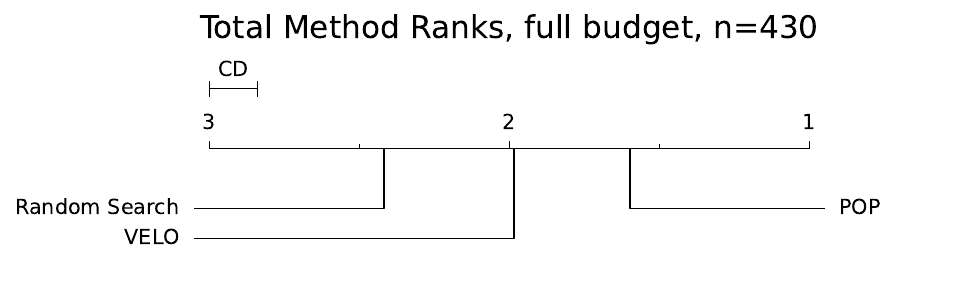}
        \caption{POP vs. L2O}
        \label{subfig:pop_vs_l2O}
    \end{subfigure}\\
    \begin{subfigure}{\textwidth}
        \centering
        \includegraphics[width=0.6\textwidth, trim={0 0 0 1cm}, clip]{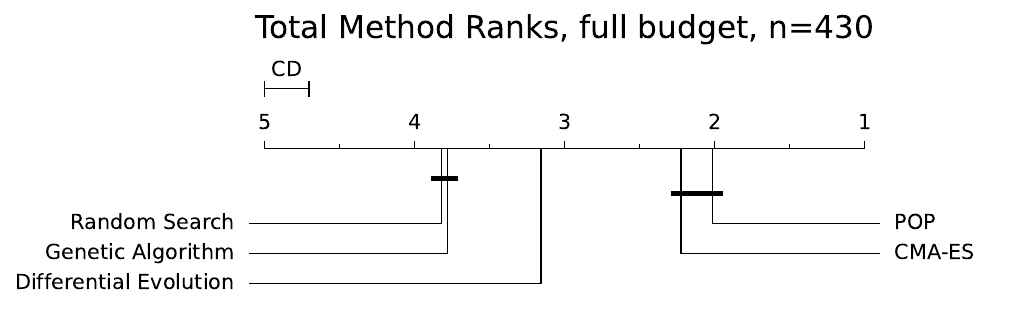}
        \caption{POP vs. Evolutionary Methods}
        \label{subfig:pop_vs_ES}
    \end{subfigure}\\
    \caption{Performance of POP compared across different model categories and representative models.}
    \label{fig:ood_categories_2}
\end{figure}

\Cref{fig:ood_categories_2} shows a more detailed comparison of POP across different model families and their representative methods.
In the out-of-distribution setting, POP performs comparably to the GP-based Bayesian optimization family, with no statistically significant difference (cf. \Cref{subfig:pop_vs_BO}). Notably, this competitive performance comes at a substantially lower computational cost: for 100 optimization iterations on a 6D problem, POP requires only 0.828 seconds, compared with 15 seconds for GP and 140.8 seconds for HEBO, corresponding to speedups of approximately 18× and 170×, respectively. POP further shows superior performance and achieves the lowest rank among the other four paradigms: first-order methods (cf. \Cref{subfig:pop_vs_first}), second-order methods (cf. \Cref{subfig:pop_vs_second}), learned optimizers (L2O; cf. \Cref{subfig:pop_vs_l2O}), and evolutionary strategies (cf. \Cref{subfig:pop_vs_ES}).
